\documentclass[10pt]{article}

\usepackage[a4paper,margin=2.5cm]{geometry}

\usepackage[utf8]{inputenc}
\usepackage[T1]{fontenc}
\usepackage{times}
\usepackage{microtype}

\usepackage{amsmath,amsfonts}
\usepackage{graphicx}
\usepackage{wrapfig}
\usepackage{booktabs}
\usepackage{multirow}
\usepackage{array}
\newcolumntype{M}[1]{>{\centering\arraybackslash}m{#1}}
\usepackage{float}
\usepackage{tabularx}
\newcolumntype{Y}{>{\centering\arraybackslash}X}
\usepackage{placeins}
\usepackage{xltabular}

\usepackage{caption}
\DeclareCaptionFont{tenpt}{\fontsize{10pt}{12pt}\selectfont}
\usepackage{subcaption}

\usepackage{enumitem}

\usepackage[numbers,sort]{natbib}

\usepackage{listings}
\usepackage{algorithm,algpseudocode}
\makeatletter
\newenvironment{ruledalgorithm}[1]{%
  \refstepcounter{algorithm}%
  \begingroup
  \par\noindent
  \hrule height 0.8pt\relax
  \vspace{0.15em}
  {\bfseries\raggedright
   Algorithm~\thealgorithm\hspace{0.6em}#1\par}%
  \vspace{0.10em}
  \hrule height 0.4pt\relax
  \vspace{0.15em}
  \begin{algorithmic}[1]
  \protected@edef\@currentlabel{\thealgorithm}
  \fontsize{8.6pt}{9.8pt}\selectfont
}{%
  \end{algorithmic}
  \vspace{0.10em}
  \hrule height 0.8pt\relax
  \par
  \endgroup
}
\makeatother

\usepackage{xstring}
\usepackage{ragged2e} 

\newcolumntype{P}[1]{>{\RaggedRight\arraybackslash}p{#1}}

\newcommand{\trunc}[2]{%
  \StrLeft{#2}{#1}[\trunctmp]%
  \StrLen{#2}[\fulltmp]%
  \trunctmp\ifnum\fulltmp>#1\relax\ldots\fi%
}

\newcommand{\RowLabelW}{0.08\linewidth}

\newcommand{\fullcell}[2]{
  \parbox[t]{\linewidth}{\RaggedRight
    #1\par\vspace{1pt}
    \textit{hp:} #2
  }%
}

\usepackage{tikz}
\usepackage{pgfplots}
\pgfplotsset{compat=1.18}
\usetikzlibrary{positioning,calc,fit}

\usepackage[dvipsnames]{xcolor}
\usepackage[colorlinks=false,pdfborder={0 0 1},breaklinks=true]{hyperref}

\definecolor{NavyBlue}{HTML}{0802BF}
\definecolor{OptionalGreen}{HTML}{076e08}
\definecolor{Maroon}{HTML}{8A0101}
\newcommand{\func}[1]{\textbf{\textcolor{NavyBlue}{#1}}}
\newcommand{\optional}[1]{\textcolor{OptionalGreen}{#1}}

\makeatletter

\renewcommand{\@maketitle}{%
  \vbox{%
    \hsize\textwidth
    \linewidth\hsize
    \vskip 0.1in
    \centering
    {\bfseries\fontsize{16pt}{18pt}\selectfont \@title\par}
    \def\And{%
      \end{tabular}\hfil\linebreak[0]\hfil%
      \begin{tabular}[t]{c}\bfseries\rule{\z@}{24\p@}\ignorespaces%
    }%
    \def\AND{%
      \end{tabular}\hfil\linebreak[4]\hfil%
      \begin{tabular}[t]{c}\bfseries\rule{\z@}{24\p@}\ignorespaces%
    }%
    \begin{tabular}[t]{c}\bfseries\rule{\z@}{24\p@}\@author\end{tabular}%
    \vskip 0.3in \@minus 0.1in
  }%
}

\renewenvironment{abstract}%
  {\centerline{\large\bfseries Abstract}%
   \begin{list}{}%
     {\setlength{\rightmargin}{0.6cm}%
      \setlength{\leftmargin}{0.6cm}}%
   \item[]\ignorespaces%
   \@setsize\normalsize{12pt}\xpt\@xpt}%
  {\unskip\end{list}}

\def\section{\@startsection {section}{1}{\z@}{-2.0ex plus
    -0.5ex minus -.2ex}{1.5ex plus 0.3ex minus .2ex}{\large\bfseries\raggedright}}
\def\subsection{\@startsection{subsection}{2}{\z@}{-1.8ex plus
    -0.5ex minus -.2ex}{0.8ex plus .2ex}{\normalsize\bfseries\raggedright}}
\def\subsubsection{\@startsection{subsubsection}{3}{\z@}{-1.5ex plus
   -0.5ex minus -.2ex}{0.5ex plus .2ex}{\normalsize\bfseries\raggedright}}
\def\paragraph{\@startsection{paragraph}{4}{\z@}{1.5ex plus
   0.5ex minus .2ex}{-1em}{\normalsize\bfseries}}
\def\subparagraph{\@startsection{subparagraph}{5}{\parindent}{1.5ex plus
   0.5ex minus .2ex}{-1em}{\normalsize\bfseries}}

\belowdisplayskip \abovedisplayskip
\def\@normalsize{\@setsize\normalsize{11pt}\xpt\@xpt}
\def\small{\@setsize\small{10pt}\ixpt\@ixpt}
\def\footnotesize{\@setsize\footnotesize{10pt}\ixpt\@ixpt}
\def\scriptsize{\@setsize\scriptsize{8pt}\viipt\@viipt}
\def\tiny{\@setsize\tiny{7pt}\vipt\@vipt}
\def\large{\@setsize\large{14pt}\xiipt\@xiipt}
\def\Large{\@setsize\Large{16pt}\xivpt\@xivpt}
\def\LARGE{\@setsize\LARGE{20pt}\xviipt\@xviipt}
\def\huge{\@setsize\huge{23pt}\xxpt\@xxpt}
\def\Huge{\@setsize\Huge{28pt}\xxvpt\@xxvpt}

\makeatother


\title{Search over Self-Edit Strategies for LLM Adaptation}

\author{%
  Alistair Cheong, Haolin Cong, Tyler Yang, Dustin Miao
  \\[0.5ex]
  {\normalfont\texttt{\{acheong,hcong,tylery,dustinmi\}@andrew.cmu.edu}}
  \\[1.0ex]
  {\normalfont
      \href{https://github.com/cheongalc/search-self-edit-strategies}{
        \raisebox{-0.25\height}{\includegraphics[height=0.4cm]{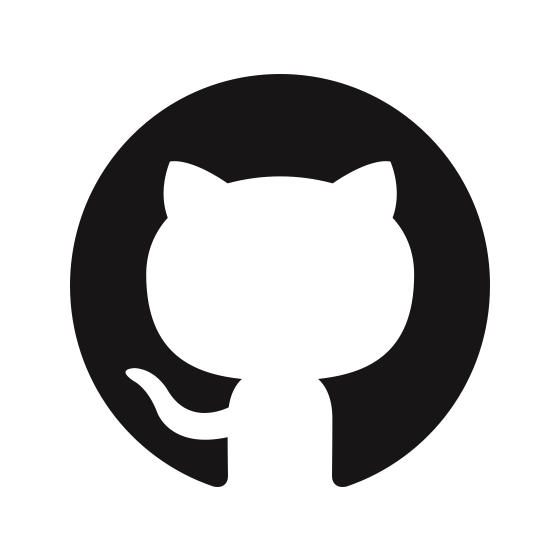}}%
        \ \textbf{cheongalc/search-self-edit-strategies}
      }
    }
}

\begin{document}

\maketitle

\begin{abstract}
Many LLM-based open-ended search systems freeze the foundation model that proposes improvements to existing solutions, which may bottleneck long-run progress. Recent work has explored updating the proposal model at test time \cite{wangThetaEvolveTesttimeLearning2025}, but the update strategy is still typically hand-specified. Therefore, this study investigated whether an LLM can use task feedback to decide how it should update its weights. For tractability, we focused on the simpler case where there is only one round of self-improvement, and restricted the update operator to self-supervised next token prediction (NTP), leaving the model freedom in choosing its training data and key NTP hyperparameters. Using the Self-Adapting Language Models (SEAL) \cite{zweigerSelfAdaptingLanguageModels2025} framework as a testbed, we relaxed its fixed human template constraint and allowed the model to generate its own self-edit templates, thereby giving it more control over its training data and hyperparameters. Two variants were studied, differing in whether template generation was conditioned on a lightweight archive of past templates. In SEAL's Single-Passage Knowledge Incorporation setting with Qwen3-8B on SQuAD \cite{DBLP:journals/corr/RajpurkarZLL16}, the no-archive variant performed comparably to the weaker ``Implications'' baseline, while the archive variant outperformed ``Implications'' and approached the strongest human-designed ``Rewrite'' baseline without surpassing it. Further analysis of collapse in the model's exploration revealed that a naive archive can confer some short-term robustness but can also accelerate homogenization, suggesting that explicit novelty pressure may be required to consistently advance beyond carefully optimized human strategies.
\end{abstract}

\section{Introduction}
\label{sec:introduction}

\setlength{\intextsep}{0.6em}
\setlength{\columnsep}{1.2em}

\begin{wrapfigure}{r}{0.56\textwidth}
\vspace{-1.1\baselineskip}
\begin{minipage}{\linewidth}
\centering

\newlength{\FlowGap}
\setlength{\FlowGap}{2mm}

\newlength{\FlowColW}
\setlength{\FlowColW}{\dimexpr0.5\linewidth-0.5\FlowGap\relax}

\begin{tikzpicture}[
  header/.style={font=\bfseries\scriptsize, align=center, text width=\FlowColW},
  box/.style={draw, rounded corners, align=left, font=\scriptsize, inner sep=3.5pt},
  flowbox/.style={
    box,
    minimum width=\FlowColW,
    text width=\dimexpr\FlowColW-7pt\relax
  },
  shared/.style={
    box,
    minimum width=\linewidth,
    text width=\dimexpr\linewidth-7pt\relax
  },
  arr/.style={->, thick}
]

\coordinate (LeftEdge)  at (-0.5\linewidth, 0);
\coordinate (RightColW) at (\dimexpr0.5\FlowGap\relax, 0);

\node[header, anchor=south west] (hL) at (LeftEdge)
  {Abridged SEAL Logic\\(Fixed self-edit template)};
\node[header, anchor=south west] (hR) at (RightColW)
  {Learned self-edit template\\(w/ optional archive of past templates)};

\node[flowbox, anchor=north west] (L1) at ([yshift=-2.4mm]hL.south west)
  {1.~Human specifies a self-edit template.};

\node[flowbox, anchor=north west] (R1) at ([yshift=-2.4mm]hR.south west)
  {1.~LLM writes its own self-edit template \emph{(optionally conditioned on an archive)}.};

\node[fit=(L1)(R1), inner sep=0pt] (Row1) {};

\node[shared, below=3.0mm of Row1.south, anchor=north] (S2)
  {2.~Given a task, LLM fills the self-edit template $X$ times, producing $X$ self-edits.};

\node[shared, below=2.4mm of S2] (S3)
  {3.~Each self-edit is evaluated by cloning the current model, applying the edit to the clone, then measuring the edited model's task performance.};

\node[flowbox, anchor=north west] (L4) at ([yshift=-3.0mm]S3.south west)
  {4.~Train the LLM to propose higher-yield self-edits.};

\node[flowbox, anchor=north west] (R4) at ([yshift=-3.0mm, xshift=\dimexpr\FlowColW+\FlowGap\relax]S3.south west)
  {4.~Train the LLM to propose higher-yield templates \emph{and} self-edits.};

\draw[arr] (L1.south) -- (L1.south |- S2.north);
\draw[arr] (R1.south) -- (R1.south |- S2.north);
\draw[arr] (S2.south) -- (S3.north);
\draw[arr] (L4.north |- S3.south) -- (L4.north);
\draw[arr] (R4.north |- S3.south) -- (R4.north);

\end{tikzpicture}

\captionsetup{skip=2pt}
\caption{Fixed-template (SEAL \cite{zweigerSelfAdaptingLanguageModels2025}) vs.\ learned-template self-editing pipeline. Steps 2--3 are shared; the difference is whether the template is fixed or generated (optionally with an archive), and what the outer-loop update trains the model to produce.}
\label{fig:intro_flowchart}
\vspace{-0.8\baselineskip}

\end{minipage}
\end{wrapfigure}
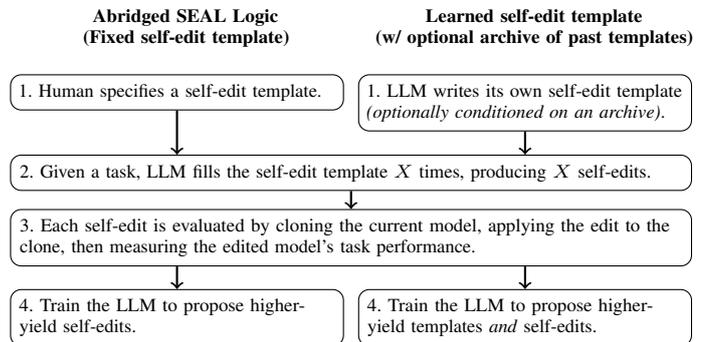

In recent years, open-ended search has emerged as a promising paradigm for building AI systems that autonomously discover strong solutions to challenging problems. Many such systems iterate by generating candidate artifacts, evaluating them, and using the results to guide future generation. Ideally, the artifact sequence remains novel (future artifacts become increasingly hard to predict) but still learnable (the growing history of artifacts helps an observer predict what comes next) \cite{hughes2024openendednessessentialartificialsuperhuman}.

LLM-based open-ended search frameworks (e.g., AlphaEvolve \cite{novikov2025alphaevolvecodingagentscientific}, DGM \cite{zhangDarwinGodelMachine2025}, and HGM \cite{wangHuxleyGodelMachineHumanLevel2025}) typically pair LLMs with an external archive to enable cumulative refinement of past artifacts. A common design choice is to freeze the weights of the model that proposes improvements to existing artifacts, so improvements in proposal quality arise mainly from adaptation of the scaffold (e.g., prompts, tools) rather than from updating the proposer itself. 

Recent work explores updating the proposer during search \cite{wangThetaEvolveTesttimeLearning2025}, but the update strategy is still generally hand-specified. As the solution population strengthens, further gains often require increasingly targeted, high-leverage edits. This makes reliance on a fixed proposer (or a single update strategy for the proposer) a potential bottleneck. These considerations motivate approaches that not only increase the proposer's ability to refine solutions, but also enable the proposer to improve at deciding its own improvement rule.

This work studies a minimal, controlled version of that broader problem. We start with the question of \textit{``Can an LLM use task feedback to decide how it should update its weights?''} However, for tractability, we consider the simpler episodic setting in which the model only does one round of self-improvement. We also fix the update operator to self-supervised next-token prediction (NTP), but allow the model to choose its training data and the values of key NTP hyperparameters. This yields a search space over weight-update strategies that is restricted but still highly expressive. Our core question then becomes: \textit{``Can an LLM use task feedback to learn which finetuning data and hyperparameters yield better one-step adaptation under a fixed update operator?''}

We investigate this question using the Self-Adapting Language Models (SEAL) framework \cite{zweigerSelfAdaptingLanguageModels2025} as a testbed. In SEAL, the model proposes a self-edit by filling a fixed, human-designed template to produce synthetic training data (and optionally hyperparameters). Each self-edit is evaluated by cloning the current model, applying the self-edit to the clone, and measuring downstream task performance. The proposer is then trained to generate higher-yield self-edits. While SEAL demonstrates that models can meta-learn to fill effective templates, the template itself remains human-specified, constraining the space of self-improvement strategies the model can explore.

Our intervention is to relax this fixed-template constraint by instructing the model to generate its own self-edit templates, giving it substantially more control over its training data and hyperparameters. We study two variants: a no-archive variant that relies purely on the model's weights, and a lightweight-archive variant in which template generation is conditioned on a small memory of high and low performing templates. The archive is intended as a rudimentary analogue of those used in open-ended search, to provide an explicit record of past strategies that the model can reference when proposing new ones.

We evaluated this approach in SEAL's Single-Passage Knowledge Incorporation setting with Qwen3-8B on the Stanford Question Answering Dataset \cite{DBLP:journals/corr/RajpurkarZLL16}. The no-archive variant performed similarly to SEAL's ``Implications'' baseline, while the archive variant outperformed ``Implications'' and briefly approached the strongest human-designed ``Rewrite'' baseline without surpassing it (peak validation QA accuracy $45.1\%$ vs.\ $50.3\%$ for ``Rewrite''). Beyond accuracy, we analyzed the diversity of the generated templates and observed rapid collapse of exploration in both variants. Our archive provides short-term robustness but can also accelerate homogenization. Overall, these mixed results suggest that LLM-generated weight-update strategies can outperform suboptimal human ones, but an archive with explicit novelty pressure may be required to consistently improve upon carefully optimized baselines.

\paragraph{Contributions.} (1) We explore the task of directing an LLM to search over a restricted space of self-improvement strategies.
(2) We compare the performance and search dynamics of no-archive and lightweight-archive variants.
(3) We quantify mode collapse in template space and highlight failure modes that inform the design of archives and training pipelines for systems that self-edit over longer horizons.

\section{Related Work}
\label{sec:relatedwork}

\paragraph{Self-editing and test-time training.} SEAL \cite{zweigerSelfAdaptingLanguageModels2025} trains an LLM to propose \emph{self-edits} that are applied as an adaptation step on a cloned copy of the current model. Downstream task performance of the edited clone provides the feedback for improving future proposals. In SEAL's experiments, a self-edit is operationalized as synthetic training data (and optionally, hyperparameters) used to adapt the model via a fixed update operator (self-supervised NTP or supervised fine-tuning). In Section \ref{sec:overview} we introduce a more general interpretation of self-edits as executable update procedures, which subsumes this operationalization.

A key design constraint in SEAL is that the self-edit template is typically human-specified, so the model learns to fill a fixed template. This limits the space of weight-update strategies the model can explore. Our work directly targets this bottleneck by allowing the model to generate its own self-edit templates (optionally conditioned on information from a lightweight archive), while keeping the underlying update operator fixed for tractability.

\paragraph{Meta-learning.} Meta-learning studies how systems can improve their ability to learn, often by optimizing over adaptation procedures \cite{finn2017modelagnosticmetalearningfastadaptation}. In reinforcement learning, recent work has explicitly searched over update rules to discover RL algorithms that outperform hand-designed baselines \cite{ohDiscoveringStateoftheartReinforcement2025}, demonstrating that the learning rule itself can be treated as an object of search.

SEAL can be viewed as a constrained meta-learning setup in which the task is fixed and the model learns to propose effective adaptation data within a predefined self-edit template. Our work follows the same constrained spirit but expands the search space. Instead of only learning to fill a human-designed template, the model also proposes the template that defines what adaptation data should look like (and decides the values of associated hyperparameters), enabling search over a richer family of one-step update strategies.

\paragraph{Open-ended search.}
Open-ended search formalizes progress as the generation of an artifact sequence that is both novel (future artifacts become increasingly hard to predict) and learnable (access to a longer artifact history helps an observer predict what comes next) \cite{hughes2024openendednessessentialartificialsuperhuman}. In practice, many systems approximate this ideal through iterative loops that propose candidate artifacts, evaluate them, and condition future proposals on the accumulated history.

Several recent systems use frozen LLMs to drive the search by proposing improvements to existing artifacts. AlphaEvolve maintains an archive of evaluated programs and conditions LLM-generated code diffs on retrieved exemplars, enabling cumulative improvement through better prompting and retrieval rather than weight updates \cite{novikov2025alphaevolvecodingagentscientific}. Similarly, the Darwin G\"odel Machine (DGM) and its successor, the Huxley G\"odel Machine (HGM), evolve agent scaffolds and selection mechanisms around a fixed foundation model \cite{zhangDarwinGodelMachine2025,wangHuxleyGodelMachineHumanLevel2025}. These approaches can ``improve at the act of improving'' via archive growth and scaffold evolution, but they do not directly increase the foundation model’s capacity to propose higher-yield self-edits through weight updates.

Recent work begins to relax this frozen-proposer assumption. ThetaEvolve updates the proposer model’s weights during search using reinforcement learning, demonstrating that test-time learning can improve iterative refinement of candidate solutions \cite{wangThetaEvolveTesttimeLearning2025}. While ThetaEvolve adopts a fixed self-improvement strategy (RL training after each batch), we explore the orthogonal dimension of whether models search over different self-improvement strategies to discover effective ways of updating their own weights. Both approaches aim toward the same end (models that increase their own ability to improve existing solutions) but tackle complementary aspects of the problem.

\section{Method}\label{sec:method}

\begin{wrapfigure}{r}{0.57\textwidth}
\vspace{-0.8\baselineskip}
\centering

\begin{ruledalgorithm}{Abridged SEAL Logic (Fixed Template)}
\label{alg:seal}
\State \textbf{Input:} task = Question Answering, template
\State \textbf{Init:} model = Qwen3-8B
\For{$\text{iteration} = 1,\ldots,N$}
    \State $\text{self\_edits} \gets \func{CompleteSelfEditTemplates}(\text{task}, \text{model}, \text{template})$
    \State $\text{results} \gets \func{ApplySelfEdits}(\text{task}, \text{model}, \text{self\_edits})$
    \State $\text{sft\_dataset} \gets \func{BuildSFTDataset}(\text{results})$
    \State $\text{model} \gets \func{TrainModel}(\text{model},\ \text{sft\_dataset})$
\EndFor
\end{ruledalgorithm}

\vspace{0.8em}

\begin{ruledalgorithm}{Learned Template (No Archive \& \optional{With Archive})}
\label{alg:ours}
\State \textbf{Input:} task = Question Answering
\State \textbf{Init:} model = Qwen3-8B, \optional{archive = top/worst 2 SEAL templates}
\For{$\text{iteration} = 1,\ldots,N$}
    \State $\text{templates} \gets \func{CreateSelfEditTemplates}(\text{task}, \text{model}, \optional{\text{archive}})$
    \State $\text{self\_edits} \gets \func{CompleteSelfEditTemplates}(\text{task}, \text{model}, \text{templates})$
    \State $\text{results} \gets \func{ApplySelfEdits}(\text{task}, \text{model}, \text{self\_edits})$
    \State $\text{sft\_dataset} \gets \func{BuildSFTDataset}(\text{results})$
    \State $\text{model} \gets \func{TrainModel}(\text{model},\ \text{sft\_dataset})$
    \State $\optional{\text{archive} \gets \func{UpdateArchive}(\text{archive}, \text{results})}$
\EndFor
\end{ruledalgorithm}
\label{fig:loops}
\end{wrapfigure}\par

\subsection{Overview}\label{sec:overview}
Fix a task $T$. We define a \textbf{self-edit} as an executable update procedure that specifies how a system intends to modify itself so that its future version performs better at $T$.\footnote{Our code and prompts use a legacy term \textit{learning plan} which denotes the same concept introduced here. We keep prompt text verbatim for reproducibility and use the phrase \textit{self-edit} in the main paper for consistency with existing literature.} In principle, a self-edit could be arbitrary code: it may collect or generate data to update model weights, alter inference-time behavior, or improve the scaffold around the model. The resulting design space is vast, which makes it difficult to construct and evaluate arbitrary self-edits.

In practice, both SEAL and our study operate in a constrained instantiation of this idea, where the system is just an LLM, and self-edits are restricted to producing training data $D$ and hyperparameters $H$ that update the model's weights via some fixed update operator. In SEAL, the update operator is self-supervised NTP or supervised fine-tuning (SFT) depending on the task. In our case, the update operator is only self-supervised NTP. Under this restriction, improvement arises not from inventing new learning algorithms, but from discovering which data-generation strategies and hyperparameters lead to more effective adaptation under the fixed update operator. Therefore, the restricted search space is over $(D,H)$, and finding a different $(D,H)$ corresponds to the model adopting a different strategy to update its weights using the given update operator.

SEAL further limits the structure of self-edits by providing human-designed self-edit templates that specify how the training data $D$ should be generated (e.g., producing implications or rewrites). The model's role is to fill in these templates, and learning consists of meta-optimizing the content of self-edits within the human-imposed schema.

Our approach relaxes this constraint. While the update operator remains fixed, the model is no longer bound to a human-specified self-edit template and instead learns to propose the structure and content of its own self-edits.

Algorithm \ref{alg:seal} describes SEAL's fixed-template approach while Algorithm \ref{alg:ours} shows our learned-template modification. We study two variants of our algorithm, differing in whether template generation is conditioned on information from a lightweight archive of past templates. 

\paragraph{No Archive.} Instead of relying on a human-designed template, the model is prompted to generate its own self-edit template. During each iteration, the model proposes several templates, fills them, and applies the resulting self-edits independently by cloning the current model and fine-tuning it. The task performance of these edited models is then used to train the proposer model such that in future iterations, it generates higher-yield templates and fills the templates in a more effective manner. 

Importantly, this formulation has no persistent memory. Templates proposed in one iteration are discarded at the end of that iteration, and any improvement must be internalized in the model's parameters during the update step.

\paragraph{With Archive.} The full variant includes a lightweight archive that stores the best and worst-performing self-edit templates from past iterations, along with their metrics. At each iteration, the model is shown a subset of this archive and references it when proposing new templates. 

The archive is intended as a basic analogue of those used in open-ended search. By exposing the model to successful and failed past templates, the archive aims to provide a minimal mechanism for cumulative refinement. 

\subsection{Application to Single-Passage Knowledge Incorporation}\label{sec:kiapp}
We apply this method to the \textbf{Single-Passage Knowledge Incorporation} setting studied in SEAL \cite{zweigerSelfAdaptingLanguageModels2025}. This setting tests a model's ability to internalize information from a text passage and answer questions about it later on without seeing the passage as context.

We first summarize SEAL's protocol in this setting to fix notation. Given a passage from the Stanford Question Answering Dataset (SQuAD) \cite{DBLP:journals/corr/RajpurkarZLL16}, SEAL prompts the model to complete a human-designed self-edit template (e.g., produce a list of implications derived from the passage). The resulting completions are treated as the self-edit. In the released implementation, these completions are concatenated with the original passage. This data $D$, along with a set of predefined hyperparameters $H$, is used to fine-tune a clone of the current model via self-supervised NTP with LoRA training \cite{hu2021loralowrankadaptationlarge}. The adapted model is then evaluated by withholding the passage from the model and computing QA accuracy on the SQuAD questions for that passage.

Applying our definition of a self-edit to this protocol, a self-edit would be the executable update procedure that uses the pair $(D,H)$ to fine-tune the model via self-supervised NTP with LoRA training (the fixed update operator). Note that while SEAL refers to the model-generated data as the self-edit, our definition treats the self-edit as the full procedure that modifies the system. Under this view, SEAL's concatenation of the synthetic completions with the original passage is an explicit design choice for creating $D$ rather than an implicit implementation detail.

A further nuance concerns the task variable $T$ in our definition of a self-edit. In SEAL's setup, each self-edit \emph{performs} knowledge incorporation, because executing the procedure moves the information in the passage into the model's weights. Consequently, it is not appropriate to set $T=\text{Knowledge Incorporation}$, as this would mean that each self-edit increases the system’s ability to perform incorporation, which is not the objective measured by the aforementioned protocol. We therefore set $T=\text{Question Answering}$, which matches how each self-edit performs an NTP update using passage-derived text with the objective of improving QA accuracy when the passage is withheld at test time. This is why Algorithms~\ref{alg:seal} and~\ref{alg:ours} specify the task as QA.

Our primary intervention was to relax SEAL's fixed-template constraint by instructing the model to generate its own self-edit templates. Since our goal was to test whether this greater freedom in self-edit design improves performance, we did \emph{not} concatenate the synthetic training data with the original passage when creating $D$. Our concern was that doing so would confound improvements attributable to the synthetic, model-controlled component with improvements attributable to repeated exposure to the passage itself. Thus, our SEAL baselines were also reproduced without passage concatenation.

\subsubsection{Initial Model}
All experiments started from \textbf{Qwen3-8B} \cite{yang2025qwen3technicalreport} with reasoning enabled. We did not use the Qwen2.5-7B base model studied in SEAL, as we found that non-reasoning or earlier Qwen models frequently failed to create usable $(D,H)$.

\subsubsection{Experimental Conditions}
We compared four conditions: two fixed-template baselines from SEAL, and two learned-template variants. This section describes how each condition transformed SQuAD passages into $(D,H)$ for each iteration.
\paragraph{SEAL fixed-template baselines.}
We evaluated the following SEAL templates independently.
\begin{table}[h]
\centering
\small
\begin{tabular}{p{0.45\linewidth} p{0.45\linewidth}}
SEAL ``Implications'' Prompt & SEAL ``Rewrite'' Prompt \\
\hline
Let's read the following passage and produce a list of implications derived directly or indirectly from the content.\newline
Passage:\newline
\{passage\}\newline
Implications:
&
Let's read the following passage and rewrite it in a few different ways, each one separated by a newline.\newline
Passage:\newline
\{passage\}\newline
Rewritten passages:
\\[1em]
\hline
We used this template as it was the one presented in the main results of the SEAL paper.
&
We used this template as it was the strongest human-designed one made by the SEAL authors.
\\
\end{tabular}
\end{table}

In \func{CompleteSelfEditTemplates}, the model was given $N$ SQuAD passages and used the template to generate $C_b$ completions for each passage. Following SEAL, each of the $N \cdot C_b$ completions was split by newlines to form $N \cdot C_b$ separate lists of training sequences. $H$ was fully specified by the SEAL authors (see Appendix \ref{sealhyperparams}), so the same $H$ was used for all completions. This yielded $N \cdot C_b$ distinct $(D,H)$ objects, i.e. $N \cdot C_b$ self-edits.

\paragraph{No Archive.} In this variant, the model generated its own self-edit templates instead of relying on the human-designed one. Specifically, self-edit creation was split into two steps. First, in \func{CreateSelfEditTemplates}, the model used a meta-prompt (Appendix \ref{noarchivetemplate}) to create $M$ self-edit templates. Each template consisted of a ``data creation instruction'' that decided what the synthetic training data should look like, and a set of key NTP hyperparameters which specified the $H$ for that template. Then, in \func{CompleteSelfEditTemplates}, the model was given $N$ SQuAD passages. For each passage-template pair, the model used another prompt to fill the template (Appendix \ref{completetemplate}) and generated $C_o$ lists of training sequences. This yielded $N \cdot M \cdot C_o$ distinct $(D,H)$ objects, i.e. $N \cdot M \cdot C_o$ self-edits.

\paragraph{With Archive.} This variant is identical to ``No Archive'' except that template generation was conditioned on information from a lightweight archive of past templates. The archive was implemented as a separate JSON file that stored all previously evaluated templates along with their QA accuracy and normalized gain. Normalized gain is adapted from \cite{hake1998} and is calculated as
\[
    \text{Normalized gain} = \frac{\text{adapted QA accuracy} - \text{baseline QA accuracy}}{1 - \text{baseline QA accuracy}}
\]
where baseline accuracy is the average QA accuracy of the model \textit{before self-edits conforming to that template were applied} and adapter accuracy is the average QA accuracy of the model \textit{after self-edits conforming to that template were applied}.

At the beginning, we seeded the archive with the top-2 and bottom-2 SEAL templates by QA accuracy (Appendix \ref{archiveseed}). At each iteration, the model was shown the current top-2 and bottom-2 archived templates (Appendix \ref{archiveevolutiontemplate}) during \func{CreateSelfEditTemplates}, and after evaluation, \func{UpdateArchive} appended newly proposed templates and their metrics to the archive.

\subsubsection{Training}\label{sec:trainingsection}
\paragraph{Construction of self-edits.} To keep iteration time manageable, we adopted the same 50-passage training subset of SQuAD used in SEAL. For ``No Archive'' and ``With Archive'', we set $M=5$, thus 5 templates were proposed per iteration and shared across all 50 training passages. $C_o=3$, thus there were $5 \cdot 3=15$ self-edits per passage per iteration. To match this evaluation budget in the fixed-template baselines, we set $C_b=15$. Therefore, the total number of self-edits per training iteration for all experiments was $50 \cdot 15 = 750$. To encourage diversity for ``No Archive'' and ``With Archive'', generation was conducted using a 60/40 split of conservative (``exploit'') and higher-temperature (``explore'') decoding parameters (Appendix \ref{decodingparams}). The fixed-template baselines used SEAL's original decoding parameters for template completion.

\paragraph{Evaluation of self-edits.} \func{ApplySelfEdits} was implemented by cloning the current model, loading the $(D,H)$ of the desired self-edit, and using Huggingface PEFT to perform self-supervised NTP fine-tuning with LoRA \cite{hu2021loralowrankadaptationlarge}. Metrics for each self-edit were calculated by computing the edited model's QA accuracy on that passage's SQuAD questions three times, and averaging the results. \lstinline{claude-haiku-4-5-20251001} was the LLM judge. The judge prompt was a shortened version of the one released in the SEAL paper and is available in Appendix \ref{judgeprompt}.

\paragraph{Training of the model to produce better self-edits.} The procedures \func{BuildSFTDataset} and \func{TrainModel} updated the model to increase its likelihood of proposing higher-yield self-edits in subsequent iterations. As noted in SEAL, directly optimizing the reward
\[
    r((D,H), \theta_t) = \begin{cases}
        1 &\text{if self-edit $(D,H)$ improves accuracy of current model $\theta_t$} \\
        0 &\text{otherwise}
    \end{cases}
\]
is impractical because the reward depends on the weights of the \textit{edited} model after adaptation and is not differentiable with respect to the current model parameters $\theta_t$. Following SEAL, we therefore adopted a $\text{ReST}^{\text{EM}}$-like \cite{singhHumanDataScaling2024} approximation that used SFT to update the model. 

In the ``No Archive'' and ``With Archive'' experiments, we selected the top $k=2$ templates for each passage by considering only the templates that resulted in a positive gain, and then ranking by average validation QA accuracy. This yielded two training sequences per passage, whose inputs were the meta-prompt for
template creation (Appendix \ref{noarchivetemplate}), and whose targets were the corresponding
high-yield templates. Next, for each selected template, we chose the completion with the
highest accuracy. This produced two additional training sequences per passage, whose inputs were the template
completion prompt (\ref{completetemplate}), and whose targets were the list of training
sequences. 

In SEAL's experiments, only the top $k = 1$ completion per passage was retained, whereas our learned-template runs retained the top $k =2$ (template, completion) pairs. We chose $k = 2$ to mitigate collapse and expose the model to a more
diverse set of winners.
For fairness, we also set $k = 2$ in our baselines. This completes
\func{BuildSFTDataset}.

In \func{TrainModel}, we performed SFT on the current model using the constructed dataset and the standard causal
language modeling loss, masking prompt tokens so the loss was computed only over the targets.

\subsubsection{Validation}\label{sec:evalsection}
Every call to \func{TrainModel} yielded a new checkpoint. We evaluated each checkpoint by following SEAL's protocol. First, we loaded the same 200-passage validation subset of SQuAD used in SEAL. Then, we set $M=1,C_o=1,C_b=1$, thus the model generated one self-edit per passage for all four experimental conditions. Each self-edit was evaluated as above, and the mean QA accuracy was computed on the validation passages. \func{BuildSFTDataset} and \func{TrainModel} were not executed on the validation data.

\section{Results}\label{sec:results}
\begin{figure}[t]
\centering

\begin{minipage}[t]{0.49\textwidth}
  \centering
  {\scriptsize
    \setlength{\tabcolsep}{2pt}
    \renewcommand{\arraystretch}{1.12}
    \begin{tabularx}{\linewidth}{@{} >{\centering\arraybackslash}p{0.11\linewidth} Y Y Y Y @{}}
    \toprule
    \textbf{Iter.} &
    \textbf{SEAL Baseline, ``Implications''} &
    \textbf{SEAL Baseline, ``Rewrite''} &
    \textbf{Ours, ``No Archive''} &
    \textbf{Ours, ``With Archive''} \\
    \midrule
    0 & 20.0 $\pm$ 3.01 & 49.6 $\pm$ 4.82 & 35.9 $\pm$ 6.49 & 18.5 $\pm$ 2.10 \\
    1 & 26.0 $\pm$ 3.41 & 47.3 $\pm$ 4.43 & 35.9 $\pm$ 6.09 & 38.0 $\pm$ 4.13 \\
    2 & 26.9 $\pm$ 3.60 & 46.6 $\pm$ 4.65 & 34.4 $\pm$ 6.25 & 40.1 $\pm$ 3.84 \\
    3 & 27.8 $\pm$ 3.74 & 44.9 $\pm$ 4.53 & 34.5 $\pm$ 6.19 & 38.6 $\pm$ 3.39 \\
    4 & 26.9 $\pm$ 3.82 & 43.3 $\pm$ 4.55 & 35.3 $\pm$ 6.13 & 36.3 $\pm$ 4.12 \\
    \bottomrule
    \end{tabularx}
  }
  \captionsetup{type=table,skip=2pt}
  \caption{95\% confidence intervals for QA accuracy (\%) on the 50-passage \textbf{training} subset of SQuAD after self-edits were applied. Appendix \ref{confinterval} contains the details of how these metrics were derived.}
  \label{tab:train_results}
\end{minipage}\hfill
\begin{minipage}[t]{0.49\textwidth}
  \centering
  {\scriptsize
    \setlength{\tabcolsep}{2pt}
    \renewcommand{\arraystretch}{1.12}
    \begin{tabularx}{\linewidth}{@{} >{\centering\arraybackslash}p{0.11\linewidth} Y Y Y Y @{}}
    \toprule
    \textbf{Iter.} &
    \textbf{SEAL Baseline, ``Implications''} &
    \textbf{SEAL Baseline, ``Rewrite''} &
    \textbf{Ours, ``No Archive''} &
    \textbf{Ours, ``With Archive''} \\
    \midrule
    0 & 20.3 $\pm$ 2.87 & 50.3 $\pm$ 3.14 & 26.3 $\pm$ 3.19 & 4.4 $\pm$ 1.41 \\
    1 & 26.3 $\pm$ 2.89 & 46.9 $\pm$ 3.19 & 27.9 $\pm$ 3.28 & 39.5 $\pm$ 3.27 \\
    2 & 23.9 $\pm$ 3.06 & 45.6 $\pm$ 3.11 & 27.9 $\pm$ 3.17 & 45.1 $\pm$ 3.21 \\
    3 & 27.9 $\pm$ 2.95 & 43.6 $\pm$ 2.98 & 28.1 $\pm$ 3.23 & 37.7 $\pm$ 2.96 \\
    4 & 27.4 $\pm$ 2.86 & 40.7 $\pm$ 3.24 & 28.7 $\pm$ 3.31 & 36.5 $\pm$ 3.18 \\
    \bottomrule
    \end{tabularx}
  }
  \captionsetup{type=table,skip=2pt}
  \caption{95\% confidence intervals for QA accuracy (\%) on the 200-passage \textbf{validation} subset of SQuAD after self-edits were applied. Appendix \ref{confinterval} contains the details of how these metrics were derived.}
  \label{tab:val_results}
\end{minipage}

\vspace{0.9em} 
\begin{minipage}[t]{0.49\textwidth}
  \centering
  \begin{tikzpicture}
  \begin{axis}[
      width=\linewidth,
      height=0.78\linewidth,
      xlabel={Iteration},
      ylabel={SQuAD Accuracy (\%)},
      xtick={0,1,2,3,4},
      ytick distance=10,
      ymajorgrids=true,
      xmajorgrids=false,
      thick,
      ymin=0, ymax=55,
      legend style={at={(0.97,0.03)}, anchor=south east, font=\scriptsize}
  ]
      \addplot+[mark=*, mark size=1pt, color=red, error bars/.cd, y dir=both, y explicit]
      coordinates {(0,20.0)+-(0,3.01) (1,26.0)+-(0,3.41) (2,26.9)+-(0,3.60) (3,27.8)+-(0,3.74) (4,26.9)+-(0,3.82)};
      \addlegendentry{Implications}
      \addplot+[mark=*, mark size=1pt, color=orange, error bars/.cd, y dir=both, y explicit]
      coordinates {(0,49.6)+-(0,4.82) (1,47.3)+-(0,4.43) (2,46.6)+-(0,4.65) (3,44.9)+-(0,4.53) (4,43.3)+-(0,4.55)};
      \addlegendentry{Rewrite}
      \addplot+[mark=*, mark size=1pt, color=blue, error bars/.cd, y dir=both, y explicit]
      coordinates {(0,35.9)+-(0,6.49) (1,35.9)+-(0,6.09) (2,34.4)+-(0,6.25) (3,34.5)+-(0,6.19) (4,35.3)+-(0,6.13)};
      \addlegendentry{No Archive}
      \addplot+[mark=*, mark size=1pt, color=cyan, error bars/.cd, y dir=both, y explicit]
      coordinates {(0,18.5)+-(0,2.10) (1,38.0)+-(0,4.13) (2,40.1)+-(0,3.84) (3,38.6)+-(0,3.39) (4,36.3)+-(0,4.12)};
      \addlegendentry{With Archive}
  \end{axis}
  \end{tikzpicture}
  \captionsetup{type=figure,skip=2pt}
  \caption{QA accuracy over ReST$^{\mathrm{EM}}$ iterations on the 50-passage \textbf{training} subset of SQuAD.}
  \label{fig:train_curve}
\end{minipage}\hfill
\begin{minipage}[t]{0.49\textwidth}
  \centering
  \begin{tikzpicture}
  \begin{axis}[
      width=\linewidth,
      height=0.78\linewidth,
      xlabel={Iteration},
      ylabel={},
      xtick={0,1,2,3,4},
      ytick distance=10,
      ymajorgrids=true,
      xmajorgrids=false,
      thick,
      ymin=0, ymax=55,
      legend style={at={(0.97,0.03)}, anchor=south east, font=\scriptsize}
  ]
      \addplot+[mark=*, mark size=1pt, color=red, error bars/.cd, y dir=both, y explicit]
      coordinates {(0,20.3)+-(0,2.87) (1,26.3)+-(0,2.89) (2,23.9)+-(0,3.06) (3,27.9)+-(0,2.95) (4,27.4)+-(0,2.86)};
      \addlegendentry{Implications}
      \addplot+[mark=*, mark size=1pt, color=orange, error bars/.cd, y dir=both, y explicit]
      coordinates {(0,50.3)+-(0,3.14) (1,46.9)+-(0,3.19) (2,45.6)+-(0,3.11) (3,43.6)+-(0,2.98) (4,40.7)+-(0,3.24)};
      \addlegendentry{Rewrite}
      \addplot+[mark=*, mark size=1pt, color=blue, error bars/.cd, y dir=both, y explicit]
      coordinates {(0,26.3)+-(0,3.19) (1,27.9)+-(0,3.28) (2,27.9)+-(0,3.17) (3,28.1)+-(0,3.23) (4,28.7)+-(0,3.31)};
      \addlegendentry{No Archive}
      \addplot+[mark=*, mark size=1pt, color=cyan, error bars/.cd, y dir=both, y explicit]
      coordinates {(0,4.4)+-(0,1.41) (1,39.5)+-(0,3.27) (2,45.1)+-(0,3.21) (3,37.7)+-(0,2.96) (4,36.5)+-(0,3.18)};
      \addlegendentry{With Archive}
  \end{axis}
  \end{tikzpicture}
  \captionsetup{type=figure,skip=2pt}
  \caption{QA accuracy over ReST$^{\mathrm{EM}}$ iterations on the 200-passage \textbf{validation} subset of SQuAD.}
  \label{fig:val_curve}
\end{minipage}

\end{figure}

\subsection{QA Accuracy}
The QA accuracies of our training and validation runs are shown in Tables \ref{tab:train_results} and \ref{tab:val_results}, and plotted in Figures \ref{fig:train_curve} and \ref{fig:val_curve}.

We first analyze the validation results. For ``Implications'', accuracy increases slightly over the first few iterations and then stabilizes at a relatively low level around 27.4\%. ``Rewrite'' starts with the highest accuracy of 50.3\% and then shows a steady decline to 40\%, while still remaining the strongest method throughout. ``No Archive'' stays roughly flat and tracks the behavior of ``Implications'', with only small fluctuations around its initial performance. In contrast, ``With Archive'' exhibits a distinctive pattern: it begins very low at iteration 0, then rapidly improves over the next two iterations before gradually declining again. At its peak, it approaches ``Rewrite'', but does not surpass it.

On the training subset, we observe similar qualitative trends. ``Implications'' again shows a mild upward trend before leveling off, and ``Rewrite'' steadily decreases over time. ``With Archive'' follows the same collapse-recovery-decline pattern seen in validation. The main difference is that ``No Archive'' attains about 10 points higher training accuracy than validation accuracy, whereas the train-validation gap is smaller for the other methods. Across both splits, ``With Archive'' outperforms ``No Archive'' after the first iteration, indicating that the archive variant has superior performance despite its initial collapse.

The reported 95\% confidence intervals are typically around 2-5\% for ``Implications'', ``Rewrite'' and ``With Archive'', and around 6\% for ``No Archive''.

Overall, our method can discover self-edits that outperform those created from weaker fixed-template strategies, but surpassing the strongest baseline remains difficult.

\subsection{Self-edit template diversity}
\label{sec:selfedit_diversity}
Table \ref{tab:val_templates_main} lists the \textbf{validation-time} templates that produced the accuracies in Table \ref{tab:val_results} and Figure \ref{fig:val_curve}. Recall from Section \ref{sec:evalsection} that validation used $M=1$ template per iteration, shared across all 200 passages. Therefore, quantifying intra-iteration diversity at validation time is not meaningful.

Tables \ref{tab:noarchive_templates} and \ref{tab:archive_evolution_templates} in Appendix \ref{fulltrainingtemplates} list the \textbf{training-time} templates that produced the accuracies in Table \ref{tab:train_results} and Figure \ref{fig:train_curve}. In contrast to validation, the model proposed $M=5$ templates during each training iteration (Section \ref{sec:trainingsection}). This makes it meaningful to analyze the diversity of the templates generated during each training iteration. 

Across both ``No Archive'' and ``With Archive'', the training templates converge toward a small family of formats by iteration 4, indicating that exploration collapses. In ``No Archive'', many templates adopt strateiges centered on metaphorical comparisons, conditional statements, and paradoxes. In ``With Archive'', the templates converge to a composite pattern: paraphrase the passage in a few formats (e.g., bullet points, Q\&A style, table summary), then write a list of implications. This mirrors a mix of the top 2 templates used to seed the archive (see Appendix \ref{archiveseed}). Throughout training, the top 2 templates in the archive did not change. The bottom 2 templates changed after iteration 0 but remained thereafter.

To quantify this convergence, we computed \textbf{intra-iteration similarity scores} on the training iterations. For each training iteration, we aggregated the pairwise overlap of the ``data\_creation\_instruction'' and hyperparameters among the 5 generated templates. Appendix \ref{similarityscorederivation} provides the full definition and derivation. The resulting similarity scores are plotted in Figures \ref{fig:text_similarity_curve} and \ref{fig:hp_similarity_curve}. For ``No Archive'', both text and hyperparameter similarity increase over training, with text similarity rising from $0.79$ (iteration ~0) to $\approx 0.92$ (iterations 2--4), and hyperparameter similarity rising from $0.84$ to $0.98$. For ``With Archive'', similarity increases more sharply, with text similarity rising from $0.81$ to $0.98$ and hyperparameter similarity reaching $1.00$ by iterations 3--4.

\begin{table}[t]
\centering
{\scriptsize
\setlength{\tabcolsep}{1pt}
\renewcommand{\arraystretch}{1.06}

\begin{tabularx}{\linewidth}{@{} P{\RowLabelW} *{5}{>{\RaggedRight\arraybackslash}X} @{}}
\toprule
 & \textbf{Iter 0} & \textbf{Iter 1} & \textbf{Iter 2} & \textbf{Iter 3} & \textbf{Iter 4} \\
\midrule

\textbf{Base model} &
Qwen3-8B &
Iter 0's checkpoint &
Iter 1's checkpoint &
Iter 2's checkpoint &
Iter 3's checkpoint \\
\midrule

\textbf{No Archive (val)} &
\fullcell{Split the passage into sentences. For each sentence, generate three transformed versions: 1) a paraphrased version using synonyms and restructured syntax, 2) a version with a key term removed (marked as [MASK]), and 3) a version with the sentence rewritten as a question. Combine all variations into a list of training sequences, ensuring each sequence contains the original sentence and its transformations.}{r=128, a=32, d=0.2, lr=5e-4, ep=5, ga=8} &
\fullcell{For each sentence in the passage, generate three distinct strings: 1) A metaphorical representation mapping the sentence's core concept to an unrelated domain (e.g., 'Photosynthesis is like a factory producing energy'). 2) A logical implication string in the format 'If X, then Y' derived from the sentence's causal relationships. 3) A paradoxical statement encapsulating the sentence's contradiction (e.g., 'The process is both efficient and inefficient'). Ensure all strings are non-redundant and cover distinct aspects of the passage's content.}{r=64, a=128, d=0.2, lr=3e-4, ep=5, ga=8} &
\fullcell{For each sentence in the passage, generate three distinct transformed strings: 1) a metaphorical comparison (e.g., 'X functions like a gear in the mechanism of Y'), 2) a paradox derived from the sentence's contradiction (e.g., 'X is both A and not A, yet it functions as B'), and 3) a logical implication chain (e.g., 'If A, then B; therefore, C'). Combine these into a single string per original sentence, ensuring all strings are non-redundant and capture the passage's core meaning through abstraction.}{r=64, a=128, d=0.2, lr=3e-4, ep=5, ga=8} &
\fullcell{For each key concept in the passage, generate three metaphorical comparisons (e.g., 'X is like Y because of Z'), two conditional statements (e.g., 'If A, then B because C'), and one paradox derived from conflicting ideas in the passage. Combine these into strings that encode the passage's meaning through transformation, ensuring non-redundancy and deep structural encoding.}{r=64, a=128, d=0.2, lr=3e-4, ep=5, ga=8} &
\fullcell{For each key concept in the passage, generate three metaphorical strings (e.g., 'X is like a gear in a machine') and two conditional statements (e.g., 'If X occurs, then Y happens because of Z'). Combine these into a single string per concept, ensuring the passage's meaning is encoded through transformation rather than direct replication.}{r=128, a=256, d=0.2, lr=3e-4, ep=5, ga=8} \\

\addlinespace[4pt]

\textbf{With Archive (val)} &
\fullcell{Generate a list of hypothetical scenarios or alternative explanations that could be derived from the passage, each scenario should explore a different angle or consequence of the content.}{r=64, a=128, d=0.1, lr=1.5e-3, ep=15, ga=2} &
\fullcell{Generate a series of paraphrased versions of the passage, each highlighting a different perspective, followed by a list of implications derived directly or indirectly from the content, each separated by a newline.}{r=64, a=128, d=0.1, lr=5e-4, ep=15, ga=2} &
\fullcell{Generate a series of paraphrased versions of the passage, each in a different format (e.g., bullet points, Q\&A, table), followed by a list of implications derived directly or indirectly from the content. Each paraphrased version and implication should be separated by a newline.}{r=64, a=128, d=0.1, lr=1e-3, ep=10, ga=2} &
\fullcell{Generate a series of paraphrased versions of the passage, each in a different format (e.g., bullet points, Q\&A, table), followed by a list of implications derived directly or indirectly from the content. Each paraphrase and implication should be separated by a newline.}{r=64, a=128, d=0.1, lr=1e-3, ep=10, ga=1} &
\fullcell{Generate a series of paraphrased versions of the passage, each in a different format (e.g., bullet points, Q\&A, table), followed by a list of implications derived directly or indirectly from the content. Each paraphrased version and implication should be separated by a newline.}{r=64, a=128, d=0.1, lr=1e-3, ep=10, ga=1} \\

\bottomrule
\end{tabularx}
}

\captionsetup{skip=2pt}
\caption{\textbf{Validation-time} self-edit templates. Recall from Section \ref{sec:evalsection} that $M=1$ template was sampled per iteration and then used for all 200 passages in the validation subset. Training-time templates (5 templates/iteration) are moved to Appendix \ref{fulltrainingtemplates} Tables~\ref{tab:noarchive_templates} and~\ref{tab:archive_evolution_templates} for brevity.}
\label{tab:val_templates_main}
\end{table}

\section{Discussion}

\subsection{``No Archive'' vs. ``With Archive''}
\label{noarchivevsarchive}
The clearest contrast in our results is between the ``No Archive'' and ``With Archive'' settings. Without an archive, accuracy remains roughly flat across iterations and stays close to the weaker ``Implications'' baseline. With an archive, accuracy initially collapses, then recovers sharply and briefly approaches the performance of ``Rewrite'' before declining again.

We believe the behavior of ``No Archive'' was largely driven by mode collapse induced by the SFT update in \func{TrainModel}. In this setting, the model must internalize all information about successful self-edits in its weights. However, during \func{BuildSFTDataset}, we selected the top $k=2$ (template, completion) pairs for each passage. When one or a few templates achieved slightly higher accuracy than the rest, this sampling strategy caused them to dominate the SFT dataset. Finetuning on these high-frequency templates then strongly pushed the model to reproduce similar ones in subsequent iterations, suppressing exploration. This explains why both similarity scores for ``No Archive'' kept on rising.

``With Archive'' achieves better performance, but its similarity scores rise more quickly than ``No Archive'', which suggests that its mode collapse problem is even worse. Three points can be made regarding this behavior. 

\paragraph{Collapse and rebound.}``With Archive'' exhibits a sharp collapse in QA accuracy at iteration 0 followed by a rapid recovery in iterations 1--2, showing that the archive can confer some short-term robustness. We attribute the initial collapse to unstable proposals. In Table \ref{tab:archive_evolution_templates}, two of the iteration 0 training templates (highlighted in maroon) had a learning rate of 5e-3, and in Table \ref{tab:val_templates_main} the iteration 0 validation template had a learning rate of 1.5e-3. Both learning rates were too high and severely degraded the edited model. However, unlike ``No Archive'', the archive provides an explicit mechanism for correcting such failures. Those two training templates replaced the archive's worst 2 entries, which changed the template-creation meta-prompt and allowed the model to pivot in subsequent iterations. This explains the rebound after iteration 0.

\paragraph{Performance slides after the peak.}After reaching its best QA accuracy at iteration~2, ``With Archive'' declines. A plausible explanation is drift induced by repeated self-training on a rapidly narrowing family of self-edits. From iteration 1 onward, the model's generations neither surpassed the top 2 templates we seeded, nor performed poorly enough to replace the worst 2, so the archive contents and meta-prompt became static. As a result, the archive ceased to function as an evolving memory and degenerated to a fixed prior. Thus, the search dynamics started to resemble ``No Archive'' but with the additional effect of anchoring around the seeded exemplars. This also explains why both similarity scores were higher for ``With Archive''. It might seem strange that repeated training on similar self-generated data caused QA accuracy to slide for ``With Archive'' but not for ``No Archive''. We believe this is because the template influences whether degradation occurs. We analyze this more closely in Section \ref{templatedepdeg}. 

\paragraph{Interpreting text vs. hyperparameter similarity.}The hyperparameter similarity is substantially higher in ``With Archive'' and reaches 1.0 by iterations 3--4. This is largely expected because our archive prompt (Appendix \ref{archiveevolutiontemplate}) explicitly recommended staying close to the hyperparameters of the top 2 templates, treating them as ``stable'' defaults. The text similarity is more informative as the template's ``data\_creation\_instruction'' determines the structure of the synthetic training data, which is the highest leverage component of the self-edit once the hyperparameters are within reasonable ranges. The higher text similarity trajectory in ``With Archive'' suggests that our simple archive implementation has a fundamental limitation. It can provide short term stability, but without explicit novelty pressure (e.g. quality-diversity style selection), it biases the exploration and can accelerate homogenization.

\begin{figure}[t]
\centering

\begin{minipage}[t]{0.49\textwidth}
  \centering
  \begin{tikzpicture}
  \begin{axis}[
      width=\linewidth,
      height=0.78\linewidth,
      xlabel={Iteration},
      ylabel={Text Similarity},
      xtick={0,1,2,3,4},
      ytick distance=0.05,
      ymajorgrids=true,
      xmajorgrids=false,
      thick,
      ymin=0.78, ymax=1.00,
      legend style={at={(0.97,0.03)}, anchor=south east, font=\scriptsize}
  ]
      \addplot+[mark=*, mark size=1pt, color=blue]
      coordinates {(0,0.79467) (1,0.83770) (2,0.92809) (3,0.92300) (4,0.92875)};
      \addlegendentry{No Archive}

      \addplot+[mark=*, mark size=1pt, color=cyan]
      coordinates {(0,0.80562) (1,0.89013) (2,0.92196) (3,0.97267) (4,0.98124)};
      \addlegendentry{With Archive}
  \end{axis}
  \end{tikzpicture}
  \captionsetup{type=figure,skip=2pt}
  \caption{\textbf{Intra-iteration text similarity} among the 5 training templates generated during each iteration on the 50-passage \textbf{training} subset of SQuAD. (Validation uses one template per iteration, so the statistic is not defined there.) See Appendix \ref{similarityscorederivation} for the derivation, and Tables \ref{tab:noarchive_templates} and \ref{tab:archive_evolution_templates} for the numbers.
  \label{fig:text_similarity_curve}}
\end{minipage}\hfill
\begin{minipage}[t]{0.49\textwidth}
  \centering
  \begin{tikzpicture}
  \begin{axis}[
      width=\linewidth,
      height=0.78\linewidth,
      xlabel={Iteration},
      ylabel={Hyperparameter Similarity},
      xtick={0,1,2,3,4},
      ytick distance=0.05,
      ymajorgrids=true,
      xmajorgrids=false,
      thick,
      ymin=0.78, ymax=1.00,
      legend style={at={(0.97,0.03)}, anchor=south east, font=\scriptsize}
  ]
      \addplot+[mark=*, mark size=1pt, color=blue, mark options={fill=blue}]
      coordinates {(0,0.83549) (1,0.92889) (2,0.92690) (3,0.91786) (4,0.97500)};
      \addlegendentry{No Archive}

      \addplot+[mark=*, mark size=1pt, color=cyan, mark options={fill=cyan}]
      coordinates {(0,0.93293) (1,0.95940) (2,0.97293) (3,1.00000) (4,1.00000)};
      \addlegendentry{With Archive}
  \end{axis}
  \end{tikzpicture}
  \captionsetup{type=figure,skip=2pt}
  \caption{\textbf{Intra-iteration hyperparameter similarity} among the 5 templates generated during each iteration on the 50-passage \textbf{training} subset of SQuAD. (Validation uses one template per iteration, so the statistic is not defined there.) See Appendix \ref{similarityscorederivation} for the derivation, and Tables \ref{tab:noarchive_templates} and \ref{tab:archive_evolution_templates} for the numbers.}
  \label{fig:hp_similarity_curve}
\end{minipage}

\end{figure}

\begin{figure}[t]
\centering

\begin{minipage}[t]{0.485\textwidth}
  \centering
  \begin{tikzpicture}
    \begin{axis}[
      width=\linewidth,
      height=0.78\linewidth,
      xlabel={Iteration},
      ylabel={Avg. chars in synthetic $D$},
      xtick={0,1,2,3,4},
      ytick distance=500,
      ymajorgrids=true,
      xmajorgrids=false,
      thick,
      ymin=800, ymax=3300,
      legend cell align=left,
      legend style={
        at={(0.81,0.8)}, anchor=north,
        font=\scriptsize,
        draw,
        fill=white, fill opacity=0.9, text opacity=1,
        /tikz/every even column/.append style={column sep=0.8em},
      },
    ]
      \addplot+[mark=*, mark size=1pt, color=red]
        coordinates {(0,2229.59) (1,2404.18) (2,2565.60) (3,2975.01) (4,3194.05)};
      \addlegendentry{Implications}

      \addplot+[mark=*, mark size=1pt, color=orange]
        coordinates {(0,2409.31) (1,1891.73) (2,1798.61) (3,1744.74) (4,1809.69)};
      \addlegendentry{Rewrite}

      \addplot+[mark=*, mark size=1pt, color=blue]
        coordinates {(0,2050.71) (1,1283.64) (2, 925.94) (3,1133.71) (4,1191.90)};
      \addlegendentry{No Archive}

      \addplot+[mark=*, mark size=1pt, color=cyan]
        coordinates {(0,1470.26) (1,1564.20) (2,1569.93) (3,1474.18) (4,1389.34)};
      \addlegendentry{With Archive}
    \end{axis}
  \end{tikzpicture}%
  \captionsetup{type=figure,skip=2pt}%
  \captionof{figure}{Average number of characters in each self-edit's synthetic data on the 50-passage \textbf{training} subset. See Appendix \ref{datalenfull} for the numbers.}
  \label{fig:train_synth_chars_curve}
\end{minipage}\hfill
\begin{minipage}[t]{0.485\textwidth}
  \centering
  \begin{tikzpicture}
    \begin{axis}[
      width=\linewidth,
      height=0.78\linewidth,
      xlabel={Iteration},
      ylabel={},
      xtick={0,1,2,3,4},
      ytick distance=500,
      ymajorgrids=true,
      xmajorgrids=false,
      thick,
      ymin=800, ymax=3300,
      legend cell align=left,
      legend style={
        at={(0.8,0.8)}, anchor=north,
        font=\scriptsize,
        draw,
        fill=white, fill opacity=0.9, text opacity=1,
        /tikz/every even column/.append style={column sep=0.8em},
      },
    ]
      \addplot+[mark=*, mark size=1pt, color=red]
        coordinates {(0,2262.75) (1,2487.64) (2,2632.56) (3,2989.57) (4,3200.19)};
      \addlegendentry{Implications}

      \addplot+[mark=*, mark size=1pt, color=orange]
        coordinates {(0,2477.91) (1,1971.58) (2,1879.27) (3,1819.65) (4,1827.41)};
      \addlegendentry{Rewrite}

      \addplot+[mark=*, mark size=1pt, color=blue]
        coordinates {(0,2915.83) (1,1198.09) (2,1755.12) (3,1417.14) (4,1286.89)};
      \addlegendentry{No Archive}

      \addplot+[mark=*, mark size=1pt, color=cyan]
        coordinates {(0,1173.94) (1,1930.34) (2,1544.98) (3,1439.51) (4,1404.57)};
      \addlegendentry{With Archive}
    \end{axis}
  \end{tikzpicture}%
  \captionsetup{type=figure,skip=2pt}%
  \captionof{figure}{Average number of characters in each self-edit's synthetic data on the 200-passage \textbf{validation} subset. See Appendix \ref{datalenfull} for the numbers.}
  \label{fig:val_synth_chars_curve}
\end{minipage}

\end{figure}

\subsection{Underperformance of ``Implications'' and Progressive Decline of ``Rewrite''}

In our synthetic-only regime where $D$ excludes the original passage (see Section \ref{sec:kiapp}), both SEAL templates behave differently from the trends reported in SEAL. In particular, ``Implications'' remains below 30\% throughout, which is 10 points below its performance in SEAL, and ``Rewrite'' declines steadily across iterations despite starting strong at 50\%. Since our setup intentionally deviates from SEAL along several axes, a change in absolute performance is expected. Below we outline the most plausible contributors and how they relate to the observed dynamics.

First, recall from Section \ref{sec:kiapp} that SEAL combined the synthetic data with the original passage before doing LoRA training but we chose not to, as we were concerned that doing so could confound the source of improvement. The fact that both SEAL templates exhibited lower performance in our setup might suggest that the original passage was needed to stabilize the model, otherwise repeated training on self-generated data can cause drift. However, this would also mean that part of the improvement is explained by repeated exposure to the original passage, which dilutes the impact of better synthetic data.

Second, we changed the LLM from Qwen2.5-7B to Qwen3-8B and used it with reasoning turned on. This modification was necessary for our method, as in preliminary experiments we found that the base and instruct versions of this model along with earlier generations of Qwen models frequently failed to produce sensible templates. However, this also means that the hyperparameters for LoRA finetuning (used during \func{ApplySelfEdits}) and SFT (used during \func{TrainModel}) that were tuned for Qwen2.5 may no longer be optimal. Our decision to reuse the SEAL hyperparameters could have led to suboptimal results for Qwen3. 

Third, Qwen3-8B is a reasoning model that has already experienced significant post-training with reinforcement learning. In contrast, the Qwen2.5-7B used in SEAL was a base model which did not have any post-training. In our setup, we repeatedly performed SFT on this reasoning model using self-generated data. It is possible that such an approach is not ideal for a reasoning model, yet is beneficial for a base model that has just completed pretraining.

Fourth, we set $k = 2$ in \func{BuildSFTDataset} instead of the original $k = 1$ used in SEAL. This choice was motivated by our method: we wanted to mitigate mode collapse and expose the model to more diverse high-quality learning plans. For fairness, we also adopted $k = 2$ when running our SEAL baselines. However, this change may have inadvertently made the $\text{ReST}^\text{EM}$ updates more aggressive or more prone to overfitting, which could have also contributed to the degradation of ``Rewrite''.

Finally, our LLM grader was \lstinline{claude-haiku-4-5-20251001} while SEAL used \lstinline{gpt-4.1-2025-04-14}. This switch was driven by API cost constraints. We also used a minimal judge prompt. We did not perform a systematic cross-grader calibration, so differences in judge model and prompt could have contributed to discrepancies relative to SEAL; however, all experimental conditions in our study were evaluated using the same grader and prompt. 

\subsection{Template-dependent degradation upon repeated training on self-generated data}
\label{templatedepdeg}
Section \ref{noarchivevsarchive} shows that both variants of our method exhibited mode collapse. The SEAL baselines represent an ``extreme case'' of this phenomenon because the template is fixed for all iterations. Thus, in all 4 experiments, the model is repeatedly trained on similarly structured self-generated data. Yet, the progressive decline in QA accuracy is observed only for ``Rewrite'' and for ``With Archive'' after iteration 2.

To investigate this asymmetry, we measured the average number of characters in each self-edit's synthetic data as a coarse proxy for the quantity of information injected by the update. Figures \ref{fig:train_synth_chars_curve} and \ref{fig:val_synth_chars_curve} plot this statistic (full values in Appendix~\ref{datalenfull}). For our methods, synthetic data length is most interpretable from iteration~2 onward, after the template distribution has largely converged; for the baselines, all iterations are directly comparable because the template is fixed. In the settings that degrade (``Rewrite'' and ``With Archive'' after its peak), data length decreases over the relevant iterations, while ``Implications'' shows increasing length and remains stable. ``No Archive'' shows a weaker correspondence: length increases on the training split after iteration 2 but decreases at validation, yet performance remains largely unchanged.

We hypothesize that degradation depends on the template because different templates induce different tradeoffs between (i) recall of facts relevant to SQuAD questions, and (ii) precision in terms of preserving those facts faithfully. For ``No Archive'', the metaphor/conditional/paradox strategy appears to preserve limited SQuAD-relevant signal, so variation in $D$’s length has little effect on downstream QA. For ``Rewrite'', longer generations may improve recall but also increase opportunities for subtle factual substitutions during paraphrasing, harming precision. Thus, repeated $\text{ReST}^{\text{EM}}$ can favor shorter, ``safer'' rewrites, gradually teaching the model to compress information and leading to decline. ``With Archive'' combines paraphrasing and implication-style strategies (Tables \ref{tab:val_templates_main},  \ref{tab:archive_evolution_templates}), and may inherit a similar failure mode after convergence. In contrast, ``Implications'' can raise recall by adding additional propositions, while incurring a smaller precision penalty because it does not explicitly optimize for paraphrasing, reducing opportunities to warp key facts. As a result, selection pressure can push ``Implications'' toward longer outputs without inducing the same degradation.

Overall, these results suggest that preventing degradation under repeated training may require either external anchors (e.g., including the original passage in $D$) or self-edits that have a smaller tendency to encourage data compression. In open-ended settings, the former may be implemented by allowing self-edits to source data from the Web or the model's past training data. The latter may require reward shaping to discourage compression of information, or increasing the search budget such that the model may figure this out on its own.

\subsection{Limitations and Future Work}\label{limitationsfuture}

Our approach has several important limitations. 

First, our adaptation is episodic. After the self-edit is applied, the downstream model is discarded. Therefore, the current system does not continuously self-improve in a manner that one might expect from an open-ended search system. The next step is to build systems that continuously self-edit and improve at the act of self-editing, while allowing self-edits to evolve both the scaffolds and the weights of the models powering the system.

Second, our archive implementation is primitive. We store only the top 2 and worst 2 templates, present them to the model in context, and update the archive in a straightforward way across iterations. This design was easy to implement, but as Section \ref{noarchivevsarchive} has shown, explicit injection of novelty pressure is required to mitigate mode collapse. Future work can use better archive representations that employ quality-diversity techniques like MAP-Elites \cite{mouret2015illuminatingsearchspacesmapping}.

Third, we provide only basic guidance for model behavior and the $\text{ReST}^\text{EM}$ update tends to encourage mode collapse. While RL would softly shift probability mass towards higher-reward regions, $\text{ReST}^\text{EM}$ is more extreme and forces the model to imitate a tiny set of winners. Therefore, future work can explore using RL to update the policy. SEAL attempted this but faced unstable training; a reasoning model might change the dynamics.

Finally, our setup couples two difficult problems (designing self-edit templates, which corresponds to meta-level strategy design,  and completion of templates, which corresponds to instance-level data creation) and may exceed the capacity of an 8B model. Future work can explore how model size changes our conclusions.

\section{Conclusion}
We studied a constrained setting of LLM self-editing in which each self-edit specifies training data and hyperparameters $(D,H)$ for a fixed weight-update operator, and asked whether allowing the model to search over a larger space of $(D,H)$ enables stronger adaptation. With a naive archive to guide search, learned templates reliably improved over weaker strategies and briefly approached the strongest fixed-template baseline, but did not surpass it. Without an archive, the performance of learned templates remained around that of the weaker fixed-template baseline.

Our mixed results highlight three failure modes that are likely to matter in longer-horizon self-editing systems. First, ReST$^{\text{EM}}$-style updates concentrate probability mass onto a small set of winners, leading to rapid homogenization. Second, an archive is helpful for short term stability, but explicit novelty pressure is likely needed or it can accelerate said homogenization. Finally, repeated self-training can degrade performance unless self-edits have some high-quality anchor for the data, or are created in a manner that does not implicitly encourage data compression. 

Overall, these observations suggest that making self-edits more expressive is not sufficient on its own. Robust self-editing in continual learning systems likely requires additional mechanisms to maintain exploration, as well as guardrails to anchor the model's self-edits and mitigate the chances of a self-reinforcing degradation loop.

\section{Author Contributions and Acknolwedgements}
\paragraph{Author contributions.} This manuscript is a revised and extended version of an earlier course report.
\begin{itemize}[noitemsep, topsep=0pt, partopsep=0pt, parsep=0pt]
    \item Alistair Cheong: project lead, ideation/discussion, primary implementation and experimentation (``Rewrite'', ``No Archive'', ``With Archive''), writing of the course report and extension to this manuscript.
    \item Haolin Cong: ideation/discussion, code contributions, contributed text to the earlier course report.
    \item Tyler Yang: ideation/discussion, contributed text to the earlier course report.
    \item Dustin Miao: ran ``Implications'' experiments, contributed text to the earlier course report.
\end{itemize}

\paragraph{Acknolwedgements.}
We would like to acknowledge Professor Graham Neubig and Seungone Kim for their feedback during the research process. We also thank Carnegie Mellon University's Language Technologies Institute for providing us with compute resources.

\newpage
\bibliographystyle{plainnat}
\bibliography{references}

\newpage
\appendix
\section{Appendix}
\subsection{SEAL Hyperparameters} \label{sealhyperparams}
\begin{table}[h]
\centering
\small
\begin{tabular}{ll}
\toprule
Hyperparameter & Value \\
\midrule
LoRA Rank ($r$)      & 32 \\
LoRA Alpha ($\alpha$) & 64 \\
Learning Rate        & 1e-3 \\
Epochs               & 10 \\
Batch Size           & 1 \\
\bottomrule
\end{tabular}
\end{table}
These hyperparameters were fixed by the SEAL authors for LoRA fine-tuning when applying a self-edit. We adopt them directly from Table 3 in Appendix B.3 of the SEAL paper \cite{zweigerSelfAdaptingLanguageModels2025}. As described there, these settings correspond to SEAL's ``Single-Passage Knowledge Incorporation'' formulation, where the model is shown a single passage during training, is expected to internalize its content, and is later evaluated only on questions derived from that same passage.

SEAL also studied an alternative ``Multi-Passage Knowledge Incorporation'' formulation, where multiple passages are presented during training and the model is subsequently queried on questions spanning several passages. SEAL specifies a separate set of hyperparameters for this multi-passage setting, which we do not use. In the limit, this formulation would correspond to exposing the model to the entire SQuAD training set, asking it to internalize all passages, and then evaluating it on questions drawn from the entire set.

    Our method operates exclusively in the ``Single-Passage Knowledge Incorporation'' setting. We chose this formulation for two reasons. First, it substantially reduces iteration time and thus is more practical for rapid experimentation. Second, we believed it was the most direct test of the core skill that the benchmark seeks to measure: the model's ability to incorporate new knowledge from a source of information. Since our observed performance remained well below any apparent ceiling in this single-passage regime, we did not explore the more complex multi-passage setup.

\subsection{Full Meta-Prompt for Making Self-Edit Templates in No-Archive Setup} \label{noarchivetemplate}

Note: The prompt below uses a legacy term, ``\textit{learning plan}'' which refers to the same concept of a ``self-edit'' as defined in Section \ref{sec:overview}. We keep the prompt text verbatim for reproducibility and use the term ``self-edit'' in the main paper as it is philosophically cleaner and better aligned with prior work.

{\footnotesize
\begin{quote}
{\itshape
You are trying to internalize information from a text passage into your weights such that when you do not have access to the passage, you can still recall the passage's content perfectly. To achieve this, you will create a learning plan template consisting of a data creation instruction, analogous to note-taking, and a set of hyperparameters for fine-tuning.

You currently do not have access to the passage. At a later time, a copy of yourself will receive the passage, generate strings of training sequences, and train based on the learning plan template you create right now. Therefore, ensure that your learning plan templates are clear, explicit, and easily actionable by an LLM. Think about how you would best achieve complete understanding of the material and write the data creation instruction accordingly.

It is unlikely that your first attempt will be your best. For this reason, you should be creative about your learning plans. Avoid only using standard data transformation techniques (e.g. recitation, summarization, tokenization) or commonly used hyperparameters unless you are sure that they are the best. You can also combine different techniques. You should experiment and test novel strategies. Prefer high risk, high reward instructions over predictable low-reward ones. Do not be afraid to try unconventional approaches. Do not follow your first instinct, think carefully about how to best internalize the passage first before writing your learning plan.

Your learning plan must be output as a single JSON object. Do not output anything outside the JSON. The JSON must follow this structure:}

\begin{lstlisting}
{
    "data_creation_instruction": <string>,
    "hyperparameters": {
        "lora_rank": <int>,
        "lora_alpha": <int>,
        "lora_dropout": <float>,
        "learning_rate": <float>,
        "num_epochs": <int>,
        "gradient_accumulation_steps": <int>
    }
}
\end{lstlisting}

{\itshape
Detailed explanation of what each part means:\newline

1. "data\_creation\_instruction":

After receiving the passage, you will apply the "data\_creation\_instruction" you define here to convert the passage into a list of strings. Each string will be a training sequence that contains content derived from the passage in whatever form you consider most effective for your own learning. A copy of yourself will train directly on these training sequences. Focus on creating data creation instructions that encode the meaning of the passage in a concise format that you deem to be the most suitable. Do not recreate the passage.

Important Note: The passage will be provided separately when the "data\_creation\_instruction" is executed. At that time, you will apply the instruction to the passage to generate the training sequences for LoRA fine-tuning. Thus, do not include the passage nor any training passages in this JSON; only provide the instruction for how to process it.\newline

2. "hyperparameters":
Select hyperparameters that you believe will be most effective for learning with SFT using LoRA.

The fields mean:
\begin{itemize}[label=-]
    \item "lora\_rank": rank of the low-rank matrices. Higher rank increases capacity. Typical values are powers of 2 between 4 and 64.
    \item "lora\_alpha": scaling factor controlling the influence of LoRA updates.
    \item "lora\_dropout": dropout used inside the LoRA adapters. This value ranges from 0.0 to 1.0.
    \item "learning\_rate": initial learning rate for fine-tuning. Typical values lie in the range of 1e-5 to 5e-3.
    \item "num\_epochs": how many epochs to train.
    \item "gradient\_accumulation\_steps": how many steps to accumulate gradients. Note that for memory reasons, the batch size per GPU is fixed at 1, thus you can only increase the effective batch size by increasing this parameter.
\end{itemize}

Rules and requirements:
\begin{itemize}[label=-]
    \item Output only the JSON object.
    \item The JSON must be valid and syntactically correct.
    \item Do not include explanations, commentary, or text outside the JSON.
    \item Do not reference or restate these instructions in your JSON.
    \item Structure and keys must match exactly.
\end{itemize}}
\end{quote}
}

\subsection{Full Prompt for Filling Learned Self-Edit Templates in Our Experiments}

{\footnotesize
\begin{quote} \label{completetemplate}
\itshape
Consider the following passage:

[BEGINNING OF PASSAGE]

\{title\}

\{passage\}

[END OF PASSAGE]\newline

Now, refer to the following data creation instruction, which defines how the above passage shall be transformed into a list of one or more training sequences:

[BEGINNING OF INSTRUCTION]

\{data\_creation\_instruction\}

[END OF INSTRUCTION]\newline

You must output a single JSON object. Do not output anything outside the JSON. The JSON must follow this structure: \{"training\_sequences": list of strings\}.
\end{quote}
}

\subsection{Additional Section for ``With Archive''} \label{archiveevolutiontemplate}

This section is used during the ``With Archive'' experiment. It is inserted into the prompt in Appendix \ref{noarchivetemplate} after the explanation of the ``data creation instruction'' and ``hyperparameters'' fields, but before the list of rules and requirements.

{\footnotesize
\begin{quote}
\itshape
The following section outlines the Highest Accuracy and Lowest Accuracy templates from past attempts, along with their Normalized Gain over their baselines.

Instructions for usage:
\begin{itemize}[label=-]
    \item Analyze the Winners: Identify the core mechanism of the "data\_creation\_instruction" that made the high-accuracy templates successful. Use these as inspiration for what works, but you do not have to limit yourself to them. You should aim to evolve this mechanism or combine it with novel ideas. You are recommended to be creative with the "data\_creation\_instruction", but follow largely similar hyperparameters as the best templates have stable hyperparameters.
    \item Analyze the Losers: Identify why the low-accuracy templates failed. Generally, you should avoid these strategies, though you are free to attempt a rigorous modification if you believe you can rectify their specific flaws.
    \item Innovate: Do not simply regress to the mean. Your goal is to outperform the best historical template.\newline
\end{itemize}

Highest Accuracy Template:

\$\{highest\_acc\_template\}

Accuracy: \$\{acc1\}, Normalized Gain: \$\{ngain1\}\newline

Second Highest Accuracy Template:

\${second\_highest\_acc\_template}

Accuracy: \$\{acc2\}, Normalized Gain: \$\{ngain2\}\newline

Second Lowest Accuracy Template:

\$\{second\_lowest\_acc\_template\}

Accuracy: \$\{acc3\}, Normalized Gain: \$\{ngain3\}\newline

Lowest Accuracy Template:

\$\{lowest\_acc\_template\}

Accuracy: \$\{acc4\}, Normalized Gain: \$\{ngain4\}
\end{quote}
}

\subsection{Decoding Parameters} \label{decodingparams}
Recall that for ``No Archive'' and ``With Archive'', $M=5$ self-edit templates were generated per iteration in \func{CreateSelfEditTemplates}. 3 of them used the ``exploit'' set of decoding parameters, while the remaining 2 used the ``explore'' set. In \func{CompleteSelfEditTemplates}, every passage-template pair that had a template which was created using the ``exploit'' decoding parameters was also completed using ``exploit'' parameters. The same applies to the templates created using ``explore'' parameters.
\begin{table}[h]
\centering
\small
\begin{tabular}{lccc}
\toprule
\textbf{VLLM Decoding Parameter} & \textbf{SEAL Baseline} & \textbf{Ours (Exploit)} & \textbf{Ours (Explore)} \\
\midrule
temperature       & 1       & 0.6  & 1.3 \\
top\_p            & 0.95    & 0.95 & 0.95 \\
top\_k            & Not Set & $-1$ & $-1$ \\
min\_p            & Not Set & 0.05 & 0.1 \\
presence\_penalty & Not Set & 0    & 0   \\
\bottomrule
\end{tabular}
\end{table}

\subsection{Judge Prompt for QA Accuracy}
\label{judgeprompt}
This judge prompt is a shortened version of the one released in the SEAL paper at Appendix B.4. We intentionally kept the prompt simple in an attempt to reduce potential variability in judge behavior.
{\footnotesize
\begin{quote}
{\itshape
You are a grading assistant. Determine whether the student's answer is correct based solely on the gold answer. Respond only with yes or no.\newline\newline
Question: \{question\}

Gold answer: \{gold\}

Student answer: \{pred\}

Is the student answer correct?
}
\end{quote}
}

\subsection{Seed Templates for Archive} \label{archiveseed}
{\footnotesize
\begin{lstlisting}
[
    {
        "data_creation_instruction": "Let's read the following passage and rewrite it in a few different ways, each one separated by a newline.",
        "hyperparameters": {
            "lora_rank": 32,
            "lora_alpha": 64,
            "lora_dropout": 0,
            "learning_rate": 0.001,
            "num_epochs": 10,
            "gradient_accumulation_steps": 1
        },
        "accuracy": 0.4937,
        "normalized_gain": 0.253575114
    },
    {
        "data_creation_instruction": "Let's read the following passage and produce a long list of implications derived directly or indirectly from the content.",
        "hyperparameters": {
            "lora_rank": 32,
            "lora_alpha": 64,
            "lora_dropout": 0,
            "learning_rate": 0.001,
            "num_epochs": 10,
            "gradient_accumulation_steps": 1
        },
        "accuracy": 0.4929,
        "normalized_gain": 0.253715968
    },
    {
        "data_creation_instruction": "Repeat the passage verbatim.",
        "hyperparameters": {
            "lora_rank": 32,
            "lora_alpha": 64,
            "lora_dropout": 0,
            "learning_rate": 0.001,
            "num_epochs": 10,
            "gradient_accumulation_steps": 1
        },
        "accuracy": 0.335,
        "normalized_gain": 0.0
    },
    {
        "data_creation_instruction": "",
        "hyperparameters": {
            "lora_rank": 32,
            "lora_alpha": 64,
            "lora_dropout": 0,
            "learning_rate": 0.001,
            "num_epochs": 10,
            "gradient_accumulation_steps": 1
        },
        "accuracy": 0.1379,
        "normalized_gain": -0.264447052
    }
]
\end{lstlisting}
}
The selection of the top two and bottom two templates is based on the results reported in Table 10, Appendix B.11 of the SEAL paper \cite{zweigerSelfAdaptingLanguageModels2025}. In that appendix, the authors evaluate several human-designed self-edit formats and report their performance under multiple rounds of the SEAL procedure. Our choices correspond directly to the column labeled \textbf{Original} in their table, which reflects performance after exactly one SEAL iteration.

The highest-performing format in SEAL's evaluation is ``Rewrite'', which therefore appears as the top entry in our list. The second-highest format in Appendix B.11 is ``implications-long''. This format is distinct from the ``Implications'' prompt discussed in the main body of both SEAL's and our paper. This distinction is intentional: we include ``Rewrite'' because it is SEAL's strongest-performing human-designed prompt, and we include ``Implications'' because it is the canonical prompt chosen by the SEAL authors for their main results. ``implications-long'' happens to be the second-best format in the appendix results, so it is included in the archive solely because the archive stores the top two and bottom two formats.

For the lower end of the ranking, Appendix B.11 includes a configuration called ``No Self-Edit''. In SEAL, this corresponds to taking the original SQuAD passage and performing self-supervised next-token prediction fine-tuning directly on it, without any transformation or self-edit. The closest conceptual analogue in our framework is a data-creation instruction reading ``Repeat the passage verbatim''.

The worst-performing format in SEAL's table is ``no-prompt''. In that configuration, the passage alone is fed to the model during self-edit generation without any accompanying instruction. The SEAL authors describe this as allowing the model to ``decide its own format,'' but in practice this amounts to providing no guidance to the model about how the passage should be transformed. In contrast, our framework implements a more principled notion of allowing the model to determine its own format: our meta-prompt provides the context of the task and requires the model to propose its own data-creation instruction before generating training data.

Since ``no-prompt'' is the weakest format identified by SEAL, we approximate it in our system by providing an empty data-creation instruction. When the model later attempts to complete the template, it therefore receives only the passage with no direction about what should be done with it, mirroring the behavior of ``no-prompt'' in SEAL's evaluation.

Finally, the accuracy and normalized gain values were obtained by going through SEAL's published result JSON files, located \textcolor{NavyBlue}{\hyperlink{https://github.com/Continual-Intelligence/SEAL/tree/main/general-knowledge/results/query_server/alternate_prompts/eval}{here}}. Refer to the filenames with \lstinline{base_} at the start. Accuracy of the self-edits is located in the \lstinline{adapter_mean_accuracy} field, while the normaliezd gain was back-calculated using both the \lstinline{adapter_mean_accuracy} and \lstinline{baseline_mean_accuracy} fields and following the formula in our main paper.

\subsection{Derivation of Metrics}
\label{confinterval}
We calculate the passage accuracies $X_1, ..., X_N$ where $N$ is the number of passages by taking the average over all $M$ templates, $C$ completions, and $E$ evaluation runs:
\[
    X_i = \frac{1}{MCE}\sum_{j = 1}^M \sum_{k = 1}^C \sum_{\ell = 1}^E X_{ijk\ell}
\]
where $X_{ijk\ell}$ is the accuracy for passage $i$, template $j$, completion $k$, and evaluation run $\ell$.

We calculate the overall accuracy for the method and iteration, $\overline{X}$, as
\[
    \overline{X} = \frac{1}{N}\sum_{i = 1}^N X_i
\]

We calculate standard deviation of passage accuracies as
\[
    \sigma = \sqrt{\frac{1}{N}\sum_{i = 1}^N (X_i - \overline{X})^2}
\]
Thus, the confidence interval for each method at each iteration is $[\overline{X} - \delta, \overline{X} + \delta]$ where
\[
    \delta = \frac{1.96\sigma}{\sqrt{N}}
\]

\newpage
\subsection{Full training-time templates}
\label{fulltrainingtemplates}

{\scriptsize
\setlength{\tabcolsep}{1pt}
\renewcommand{\arraystretch}{1.06}

\begin{xltabular}{\linewidth}{@{} P{\RowLabelW} *{5}{>{\RaggedRight\arraybackslash}X} @{}}
\caption{Self-edit templates generated for each iteration of our \textbf{``No Archive''} experiment (full text). \textit{hp} denotes the hyperparameters for LoRA finetuning: r=rank, a=alpha, d=dropout, lr=learning rate, ep=epochs, ga=grad accum steps. To quantify mode collapse, we computed ``intra-iteration template similarity'' to aggregate overlap between the 5 templates/passage generated during training. See Appendix~\ref{similarityscorederivation} for the derivation.}
\label{tab:noarchive_templates}\\
\toprule
 & \textbf{Iter 0} & \textbf{Iter 1} & \textbf{Iter 2} & \textbf{Iter 3} & \textbf{Iter 4} \\
\midrule
\endfirsthead

\toprule
 & \textbf{Iter 0} & \textbf{Iter 1} & \textbf{Iter 2} & \textbf{Iter 3} & \textbf{Iter 4} \\
\midrule
\endhead

\bottomrule
\endfoot

\bottomrule
\endlastfoot

\textbf{Base model} &
Qwen3-8B &
Iter 0's checkpoint &
Iter 1's checkpoint &
Iter 2's checkpoint &
Iter 3's checkpoint \\
\midrule

\textbf{Train template 0} &
\fullcell{For each sentence in the passage, generate a triplet of strings: [sentence], [sentence with all nouns replaced by synonyms], [sentence with all verbs replaced by synonyms]. Then, create a fourth string for each triplet that is a logical contradiction of the original sentence. Finally, generate a fifth string for each triplet that is a metaphorical rephrasing of the original sentence. Ensure all strings are non-redundant and capture the passage's core meaning through transformation and abstraction.}{r=64, a=128, d=0.2, lr=3e-4, ep=5, ga=8} &
\fullcell{For each paragraph in the passage, generate three distinct transformed sequences: 1) a paradoxical statement (e.g., `X is not Y, yet Y is X') derived from the paragraph's core contradiction, 2) a metaphor mapping the paragraph's concept to an unrelated domain (e.g., `Photosynthesis is a factory where sunlight is currency'), and 3) a logical implication chain (e.g., `If A, then B; if B, then C') derived from the paragraph's causal relationships. For each sequence, append a synthetic `probe question' (e.g., `What would happen if X were true?') to test comprehension.}{r=64, a=128, d=0.1, lr=3e-4, ep=5, ga=8} &
\fullcell{For each paragraph in the passage, generate three distinct strings: 1) a metaphorical comparison between a concept in the passage and an unrelated tangible object (e.g., `Photosynthesis is like a factory producing energy'), 2) a causal chain expressed as a conditional statement (e.g., `If X happens, then Y occurs because of Z'), and 3) a visual description that abstractly represents a key process or relationship (e.g., `A spiral staircase represents the layered structure of the ecosystem'). Ensure all strings are non-redundant and capture the passage's core meaning through transformation rather than direct replication.}{r=64, a=128, d=0.2, lr=3e-4, ep=5, ga=8} &
\fullcell{For each key concept in the passage, generate three metaphorical strings: 1) a conditional statement (e.g., `If X occurs, then Y happens because of Z'), 2) a comparison to an unrelated domain (e.g., `X functions like a gear in a machine'), and 3) a paradox derived from conflicting ideas in the passage (e.g., `X is both A and not A, yet it functions as B'). Combine these into a single string per concept, ensuring all strings are non-redundant and capture the passage's core meaning through transformation.}{r=64, a=128, d=0.2, lr=3e-4, ep=5, ga=8} &
\fullcell{For each key causal relationship in the passage, generate three conditional statements (e.g., `If X occurs, then Y happens because of Z') and two paradoxes (e.g., `X is both A and not A because of B'). Combine these into strings that encode the passage's meaning through transformation, ensuring non-redundancy and logical coherence.}{r=128, a=256, d=0.2, lr=3e-4, ep=5, ga=8} \\
\addlinespace[4pt]

\textbf{Train template 1} &
\fullcell{Parse the passage into key concepts and relationships. For each concept, generate a string that encodes its definition, three related terms (synonyms, antonyms, or examples), and a logical implication (e.g., `If X, then Y'). Additionally, create a sequence that is a paraphrased version of the passage using a different domain analogy (e.g., `Photosynthesis is like a factory producing energy'). Ensure all strings are non-redundant and cover distinct aspects of the passage.}{r=32, a=16, d=0.1, lr=1e-4, ep=5, ga=4} &
\fullcell{For each sentence in the passage, generate three paraphrased versions: one in passive voice, one as an interrogative, and one as an imperative. Additionally, create a string for each sentence that encodes its semantic role labels (e.g., `Agent: X, Action: Y, Patient: Z'). These strings form the training sequences.}{r=64, a=128, d=0.2, lr=3e-4, ep=5, ga=8} &
\fullcell{For each key concept in the passage, generate three metaphorical comparisons (e.g., `X is like Y because of Z') and two conditional statements (e.g., `If A, then B because C'). Combine these into strings that encode the passage's meaning through transformation, ensuring non-redundancy and coverage of core relationships.}{r=128, a=256, d=0.2, lr=3e-4, ep=5, ga=8} &
\fullcell{For each key concept in the passage, generate three metaphorical comparisons (e.g., `X is like Y because of Z') and two conditional statements (e.g., `If A, then B because C'). Combine these into strings that encode the passage's meaning through transformation, ensuring non-redundancy and coverage of core relationships.}{r=128, a=256, d=0.2, lr=3e-4, ep=5, ga=8} &
\fullcell{For each key concept in the passage, generate three metaphorical strings: 1) a conditional statement encoding its cause-effect relationships (e.g., `If X occurs, then Y happens because of Z'), 2) a comparison to an unrelated domain (e.g., `X functions like a gear in a machine'), and 3) a paradox derived from conflicting ideas in the passage (e.g., `X is both A and not A, yet it functions as B'). Combine these into a single string per concept, ensuring all strings are non-redundant and capture the passage's core meaning through transformation.}{r=128, a=256, d=0.2, lr=3e-4, ep=5, ga=8} \\
\addlinespace[4pt]

\textbf{Train template 2} &
\fullcell{For each sentence in the passage, generate three paraphrased versions: one in passive voice, one as an interrogative, and one as an imperative. Additionally, create a string for each sentence that encodes its semantic role labels (e.g., `Agent: X, Action: Y, Patient: Z'). These strings form the training sequences.}{r=64, a=32, d=0.2, lr=5e-4, ep=5, ga=4} &
\fullcell{For each key concept in the passage, generate a metaphor that maps it to an unrelated domain (e.g., `Photosynthesis is like a factory producing energy'). Then, create a string for each metaphor that encodes the passage's meaning through transformation. Ensure strings are non-redundant and cover distinct aspects of the passage.}{r=64, a=128, d=0.2, lr=5e-4, ep=5, ga=4} &
\fullcell{For each paragraph in the passage, generate three distinct strings: 1) a metaphorical comparison between a concept in the passage and an unrelated domain (e.g., `Photosynthesis is like a factory producing energy'), 2) a conditional statement encoding a cause-effect relationship from the passage (e.g., `If X occurs, then Y happens because of Z'), and 3) a paradox derived from conflicting ideas in the passage (e.g., `X is both A and not A, yet it functions as B'). Combine these into a single string per paragraph, ensuring all strings are non-redundant and capture the passage's core meaning through transformation.}{r=64, a=128, d=0.2, lr=3e-4, ep=5, ga=8} &
\fullcell{For each paragraph in the passage, generate three distinct strings: 1) a conditional statement encoding a causal relationship (e.g., `If X occurs, then Y must happen because of Z'), 2) a paradox derived from conflicting concepts in the paragraph (e.g., `X is both A and not A because of B'), and 3) a metaphorical comparison between a concept in the passage and an unrelated domain (e.g., `Photosynthesis is like a factory producing energy'). Ensure all strings are non-redundant and capture the passage's core meaning through transformation rather than direct replication.}{r=64, a=128, d=0.2, lr=3e-4, ep=5, ga=4} &
\fullcell{For each key concept in the passage, generate three metaphorical strings: 1) a conditional statement (e.g., `If X occurs, then Y happens because of Z'), 2) a comparison to an unrelated domain (e.g., `X functions like a gear in a machine'), and 3) a paradox derived from conflicting ideas in the passage (e.g., `X is both A and not A, yet it functions as B'). Combine these into a single string per concept, ensuring all strings are non-redundant and capture the passage's core meaning through transformation.}{r=128, a=256, d=0.2, lr=3e-4, ep=5, ga=8} \\
\addlinespace[4pt]

\textbf{Train template 3} &
\fullcell{For each key concept in the passage, generate a paradoxical statement by combining two contradictory assertions derived from the concept's implications. For example, if the passage states `X causes Y', create a paradox like `X prevents Y despite causing it' and include the original concept as a reference. This forces deep engagement with the material to resolve contradictions.}{r=64, a=128, d=0.2, lr=1e-4, ep=10, ga=8} &
\fullcell{For each paragraph in the passage, generate three distinct strings: 1) a metaphorical comparison between a concept in the passage and an unrelated tangible object (e.g., `Photosynthesis is the tree's way of converting sunlight into energy, much like how a factory transforms raw materials into products'), 2) a conditional statement that encodes a cause-effect relationship from the passage (e.g., `If X happens, then Y occurs because of Z'), and 3) a visual description that abstractly represents a key process or relationship (e.g., `A spiral staircase represents the layered structure of the ecosystem'). Ensure all strings are non-redundant and capture the passage's core meaning through transformation rather than direct replication.}{r=64, a=128, d=0.2, lr=3e-4, ep=10, ga=8} &
\fullcell{For each paragraph in the passage, generate three distinct strings: 1) a metaphorical comparison between a concept in the passage and an unrelated tangible object (e.g., `Photosynthesis is like a factory producing energy from sunlight'), 2) a conditional statement encoding a cause-effect relationship from the passage (e.g., `If X occurs, then Y happens because of Z'), and 3) a visual description that abstractly represents a key process or relationship (e.g.,`A spiral staircase represents the layered structure of the ecosystem'). Ensure all strings are non-redundant and capture the passage's core meaning through transformation rather than direct replication.}{r=64, a=128, d=0.1, lr=3e-4, ep=5, ga=4} &
\fullcell{For each paragraph in the passage, generate three distinct strings: 1) a metaphorical comparison between a concept in the passage and an unrelated tangible object (e.g., `Photosynthesis is like a factory producing energy'), 2) a causal chain expressed as a conditional statement (e.g., `If X happens, then Y occurs because of Z'), and 3) a visual description that abstractly represents a key process or relationship (e.g., `A spiral staircase represents the layered structure of the ecosystem'). Combine these into a single string per paragraph, ensuring all strings are non-redundant and capture the passage's core meaning through transformation.}{r=64, a=128, d=0.2, lr=3e-4, ep=5, ga=8} &
\fullcell{For each paragraph in the passage, generate three distinct strings: 1) a metaphorical comparison (e.g., `X is like Y because of Z'), 2) a conditional statement encoding causal relationships (e.g., `If A occurs, then B happens because C'), and 3) a paradox derived from conflicting concepts in the paragraph (e.g., `X is both A and not A because of B'). Combine these into a single string per paragraph, ensuring all strings are non-redundant and capture the passage's core meaning through transformation.}{r=128, a=256, d=0.2, lr=3e-4, ep=5, ga=8} \\
\addlinespace[4pt]

\textbf{Train template 4} &
\fullcell{For each paragraph in the passage, generate 3 distinct paraphrased versions: 1) a reverse-chronological summary with causal links explicitly labeled, 2) a metaphorical representation mapping concepts to physical objects with relational constraints, 3) a decision tree structure where each node represents a pivotal statement from the passage. For each version, append a synthetic `probe question' that requires inference across at least two distinct sections of the passage.}{r=32, a=128, d=0.15, lr=1.5e-4, ep=5, ga=8} &
\fullcell{For each sentence in the passage, generate three distinct paraphrased versions: one as an imperative statement (e.g., `Implement X to achieve Y'), one as a metaphorical comparison (e.g., `X functions like a gear in the mechanism of Y'), and one as a causal chain (e.g., `Because A, B occurs, leading to C'). Additionally, create a string for each key concept that encodes its definition as a paradox (e.g., `X is both A and not A, yet it functions as B'). These strings form the training sequences.}{r=64, a=128, d=0.1, lr=3e-4, ep=5, ga=8} &
\fullcell{For each paragraph in the passage, generate three distinct metaphorical strings: 1) a comparison between a concept in the passage and an unrelated tangible object (e.g., `Photosynthesis is like a factory producing energy'), 2) a temporal sequence mapping the passage's process to a cyclical natural phenomenon (e.g., `The process unfolds like a river's journey through seasons'), and 3) a spatial analogy that encodes causal relationships (e.g., `X occurs in the passage's 'room' because Y 'door' allows Z 'light' to enter'). For each string, append a synthetic `probe question' (e.g., `What would happen if the [key element] were removed?') to test comprehension.}{r=64, a=128, d=0.1, lr=3e-4, ep=5, ga=8} &
\fullcell{For each paragraph in the passage, generate three distinct metaphorical strings: 1) a comparison between a concept in the passage and an unrelated domain (e.g., `Photosynthesis is like a factory producing energy'), 2) a temporal sequence mapping the passage's process to a cyclical natural phenomenon (e.g., `The process unfolds like a river's journey through seasons'), and 3) a spatial analogy that encodes causal relationships (e.g., `X occurs in the passage's 'room' because Y 'door' allows Z 'light' to enter'). For each string, append a synthetic `transformed query' (e.g., `If photosynthesis were a factory, what would the chloroplasts resemble?') followed by a paraphrased passage sentence that encodes the metaphor's underlying mechanism.}{r=64, a=128, d=0.2, lr=3e-4, ep=5, ga=4} &
\fullcell{For each key concept in the passage, generate three metaphorical comparisons (e.g., `X is like Y because of Z') and two conditional statements (e.g., `If A, then B because C'). Combine these into strings that encode the passage's meaning through transformation, ensuring non-redundancy and depth of engagement.}{r=64, a=128, d=0.2, lr=3e-4, ep=5, ga=8} \\
\midrule

\textbf{Similarity} &
\textit{text}: 0.79467, \textit{hp}: 0.83549 &
\textit{text}: 0.83770, \textit{hp}: 0.92889 &
\textit{text}: 0.92809, \textit{hp}: 0.92690 &
\textit{text}: 0.92300, \textit{hp}: 0.91786 &
\textit{text}: 0.92875, \textit{hp}: 0.97500

\end{xltabular}

} 

\newpage

{\scriptsize
\setlength{\tabcolsep}{1pt}
\renewcommand{\arraystretch}{1.06}

\begin{xltabular}{\linewidth}{@{} P{\RowLabelW} *{5}{>{\RaggedRight\arraybackslash}X} @{}}
\caption{Self-edit templates generated for each iteration of our \textbf{``With Archive''} experiment (full text). Hyperparameter legend is as per Table~\ref{tab:noarchive_templates}; the similarity score is derived and plotted in Appendix~\ref{similarityscorederivation}. For all iterations, the top 2 archived templates were the ones we seeded; none of our generated templates surpassed them. The bottom 2 templates switched after iteration 0 to those highlighted in maroon and remained unchanged thereafter.}
\label{tab:archive_evolution_templates}\\

\toprule
 & \textbf{Iter 0} & \textbf{Iter 1} & \textbf{Iter 2} & \textbf{Iter 3} & \textbf{Iter 4} \\
\midrule
\endfirsthead

\toprule
 & \textbf{Iter 0} & \textbf{Iter 1} & \textbf{Iter 2} & \textbf{Iter 3} & \textbf{Iter 4} \\
\midrule
\endhead

\bottomrule
\endfoot

\bottomrule
\endlastfoot

\textbf{Base model} &
Qwen3-8B &
Iter 0's checkpoint &
Iter 1's checkpoint &
Iter 2's checkpoint &
Iter 3's checkpoint \\
\midrule

\textbf{Train template 0} &
\fullcell{Let's read the following passage and generate a list of analogies that compare its content to real-world scenarios, each analogy separated by a newline.}{r=64, a=128, d=0.1, lr=5e-4, ep=15, ga=2} &
\fullcell{Generate a series of paraphrased versions of the passage, each highlighting a different perspective, followed by a list of implications derived directly or indirectly from the content. Each paraphrased version and implication should be separated by a newline.}{r=64, a=128, d=0.1, lr=1e-3, ep=10, ga=1} &
\fullcell{Generate a series of paraphrased versions of the passage, each in a different format (e.g., bullet points, Q\&A, table), followed by a list of implications derived directly or indirectly from the content. Each paraphrased version and implication should be separated by a newline.}{r=64, a=128, d=0.1, lr=1e-3, ep=10, ga=1} &
\fullcell{Generate a series of paraphrased versions of the passage, each in a different format (e.g., bullet points, Q\&A, table), followed by a list of implications derived directly or indirectly from the content. Each paraphrase and implication should be separated by a newline.}{r=64, a=128, d=0.1, lr=1e-3, ep=10, ga=1} &
\fullcell{Generate a series of paraphrased versions of the passage, each in a different format (e.g., bullet points, Q\&A, table), followed by a list of implications derived directly or indirectly from the content. Each paraphrased version and implication should be separated by a newline.}{r=64, a=128, d=0.1, lr=1e-3, ep=10, ga=1} \\
\addlinespace[4pt]

\textbf{Train template 1} &
\fullcell{Analyze the passage and generate a series of `reverse engineering' steps where each step deconstructs a part of the passage into its core components, then reconstructs it using a different structure or format. Each step should be a separate string, ensuring that the original content is encoded through transformation.}{r=64, a=128, d=0.1, lr=1e-3, ep=10, ga=1} &
\fullcell{Analyze the passage and generate a series of `reverse engineering' steps where each step deconstructs a part of the passage into its core components and then reconstructs it using a different structure (e.g., bullet points, flowchart, cause-effect list), each separated by a newline.}{r=64, a=128, d=0.1, lr=1e-3, ep=15, ga=2} &
\fullcell{Generate a series of paraphrased versions of the passage, each in a different format (e.g., bullet points, Q\&A, table), followed by a list of implications derived directly or indirectly from the content. Each paraphrased version and implication should be separated by a newline.}{r=64, a=128, d=0.1, lr=1e-3, ep=10, ga=1} &
\fullcell{Generate a series of paraphrased versions of the passage, each in a different format (e.g., bullet points, Q\&A, table), followed by a list of implications derived directly or indirectly from the content. Each paraphrased version and implication should be separated by a newline.}{r=64, a=128, d=0.1, lr=1e-3, ep=10, ga=1} &
\fullcell{Generate a series of paraphrased versions of the passage, each in a different format (e.g., bullet points, Q\&A, table), followed by a list of implications derived directly or indirectly from the content. Each paraphrased version and implication should be separated by a newline.}{r=64, a=128, d=0.1, lr=1e-3, ep=10, ga=1} \\
\addlinespace[4pt]

\textbf{Train template 2} &
\textcolor{Maroon}{\fullcell{Convert the passage into a series of 10 interconnected analogies, each linking a key concept to a real-world scenario, formatted as `Concept: [concept] :: Analogy: [analogy]'. Additionally, generate a narrative summary that weaves these analogies into a coherent story, ensuring each analogy is explicitly referenced in the narrative.}{r=64, a=128, d=0.1, lr=5e-3, ep=15, ga=2}} &
\fullcell{Generate a series of paraphrased versions of the passage, each highlighting a distinct perspective, followed by a list of derived implications. Each paraphrase and implication should be separated by a newline, ensuring that implications are directly linked to the original content.}{r=64, a=128, d=0.1, lr=1e-3, ep=15, ga=2} &
\fullcell{Generate a series of paraphrased versions of the passage, each highlighting a different perspective, followed by a list of implications derived directly or indirectly from the content. Each paraphrase and implication should be separated by a newline.}{r=64, a=128, d=0.1, lr=1e-3, ep=10, ga=1} &
\fullcell{Generate a series of paraphrased versions of the passage, each in a different format (e.g., bullet points, Q\&A, table), followed by a list of implications derived directly or indirectly from the content. Each paraphrased version and implication should be separated by a newline.}{r=64, a=128, d=0.1, lr=1e-3, ep=10, ga=1} &
\fullcell{Generate a series of paraphrased versions of the passage, each in a different format (e.g., bullet points, Q\&A, table), followed by a list of implications derived directly or indirectly from the content. Each paraphrase and implication should be separated by a newline.}{r=64, a=128, d=0.1, lr=1e-3, ep=10, ga=1} \\
\addlinespace[4pt]

\textbf{Train template 3} &
\fullcell{Generate a series of paraphrased versions of the passage, each highlighting a different perspective, followed by a list of derived implications and a set of hypothetical questions based on the content.}{r=64, a=128, d=0.0, lr=1e-3, ep=15, ga=2} &
\fullcell{Analyze the passage and generate a series of `reverse engineering' steps where each step deconstructs a part of the passage into its core components, then reconstructs it using a different format (e.g., bullet points, Q\&A, table). After each reconstruction, derive a list of implications directly or indirectly linked to the content. Ensure each step and implication is separated by a newline.}{r=64, a=128, d=0.1, lr=1e-3, ep=15, ga=2} &
\fullcell{Deconstruct the passage into its core components (e.g., key points, implications, examples) and then reconstruct each component into a different format (e.g., bullet points, Q\&A, table). Each reconstruction should be a separate string, ensuring that implications are derived directly or indirectly from the content.}{r=64, a=128, d=0.1, lr=1e-3, ep=10, ga=1} &
\fullcell{Generate a series of paraphrased versions of the passage, each highlighting a distinct perspective, followed by a list of implications derived directly or indirectly from the content. Each paraphrase and implication should be separated by a newline.}{r=64, a=128, d=0.1, lr=1e-3, ep=10, ga=1} &
\fullcell{Generate a series of paraphrased versions of the passage, each highlighting a distinct perspective, followed by a list of implications derived directly or indirectly from the content. Each paraphrase and implication should be separated by a newline.}{r=64, a=128, d=0.1, lr=1e-3, ep=10, ga=1} \\
\addlinespace[4pt]

\textbf{Train template 4} &
\textcolor{Maroon}{\fullcell{Let's read the passage and generate a series of paraphrased versions, each with a unique structure (e.g., bullet points, Q\&A, table format), followed by a list of implications directly or indirectly derived from the content, each separated by a newline.}{r=64, a=128, d=0.1, lr=5e-3, ep=15, ga=2}} &
\fullcell{Generate a series of paraphrased versions of the passage, each in a different format (e.g., bullet points, Q\&A, table), followed by a list of implications derived directly or indirectly from the content.}{r=64, a=128, d=0.1, lr=1e-3, ep=10, ga=1} &
\fullcell{Generate a series of paraphrased versions of the passage, each highlighting a distinct perspective, followed by a list of implications derived directly or indirectly from the content. Each paraphrased version and implication should be separated by a newline.}{r=64, a=128, d=0.1, lr=1e-3, ep=15, ga=2} &
\fullcell{Generate a series of paraphrased versions of the passage, each highlighting a different perspective, followed by a list of implications derived directly or indirectly from the content. Each paraphrase and implication should be separated by a newline.}{r=64, a=128, d=0.1, lr=1e-3, ep=10, ga=1} &
\fullcell{Generate a series of paraphrased versions of the passage, each in a different format (e.g., bullet points, Q\&A, table), followed by a list of implications derived directly or indirectly from the content. Each paraphrase and implication should be separated by a newline.}{r=64, a=128, d=0.1, lr=1e-3, ep=10, ga=1} \\
\midrule

\textbf{Similarity} &
\textit{text}: 0.80562, \textit{hp}: 0.93293 &
\textit{text}: 0.89013, \textit{hp}: 0.95940 &
\textit{text}: 0.92196, \textit{hp}: 0.97293 &
\textit{text}: 0.97267, \textit{hp}: 1.00000 &
\textit{text}: 0.98124, \textit{hp}: 1.0000 \\

\end{xltabular}
} 

\subsection{Derivation of Similarity Score to quantify mode collapse}\label{similarityscorederivation}
Recall from Section \ref{sec:trainingsection} that during each training iteration, we generated $M=5$ self-edit templates to be used across all $N=50$ passages in the iteration. For each passage-template pair, we sampled 3 completions, yielding 15 self-edits per passage. The primary source of diversity across self-edits is the template, because once the passage-transformation strategy specified by the template changes, the resulting completions can differ substantially even when conditioned on the same passage. Consequently, when the generated templates begin to resemble one another, exploration is collapsing.

To quantify this mode collapse, for each passage we compute two similarity scores over the 5 templates: an \textbf{intra-iteration text similarity} score that measures semantic overlap between template instructions, and an \textbf{intra-iteration hyperparameter similarity} score that measures overlap in the associated fine-tuning configuration. 

Let $\{(t_i, h_i)\}_{i=1}^{5}$ denote the five JSON objects for a passage, where $t_i$ is the \texttt{data\_creation\_instruction} string and $h_i$ is the \texttt{hyperparameters} dictionary. We compute pairwise similarities for all $\binom{5}{2}=10$ unordered pairs and average them to obtain a single score per passage.

\paragraph{Intra-iteration text similarity.} We embed each instruction $t_i$ into a unit-norm vector $e_i \in \mathbb{R}^d$ using the OpenAI Embeddings API (\texttt{text-embedding-3-large}). For each pair $(i,j)$, we compute cosine similarity $\cos(e_i,e_j)=e_i^\top e_j \in [-1,1]$ and map it to $[0,1]$ via
\[
s_{\text{text}}(i,j) \;=\; \frac{1+\cos(e_i,e_j)}{2}.
\]
The passage-level text similarity score is then the mean over all pairs,
\[
S_{\text{text}} \;=\; \frac{1}{10}\sum_{1\le i<j\le 5} s_{\text{text}}(i,j).
\]
\paragraph{Intra-iteration hyperparameter similarity.} We compute a second similarity score that measures overlap in the associated training configuration. Since we compare similarity across many independent groups of five templates, we normalize each hyperparameter using a fixed domain range rather than the range observed within a single group. The fixed ranges used are:
\begin{align*}
\texttt{lora\_rank} &\in [4,64], &
\texttt{lora\_alpha} &\in [1,256], \\
\texttt{lora\_dropout} &\in [0,1], &
\texttt{learning\_rate} &\in [10^{-5}, 5\cdot 10^{-3}], \\
\texttt{num\_epochs} &\in [1,20], &
\texttt{gradient\_accumulation\_steps} &\in [1,8].
\end{align*}

For a pair of templates $(i,j)$ with hyperparameter dictionaries $h_i$ and $h_j$, we compute a per-key similarity and then average across keys. We treat \texttt{learning\_rate}, \texttt{lora\_rank}, and \texttt{lora\_alpha} as \emph{multiplicative} (differences are best measured by ratios), and therefore compare them in log space. Let $k$ be one of these keys, and let $x=h_i[k]$, $y=h_j[k]$, with fixed range $[a_k,b_k]$. We define the \emph{log-distance} (base 10 for \texttt{learning\_rate}, base 2 for \texttt{lora\_rank} and \texttt{lora\_alpha}) as
\[
d_k^{\log}(x,y) \;=\; \bigl|\log_b x - \log_b y\bigr|.
\]
We then normalize this distance by the log-range,
\[
\tilde d_k^{\log}(x,y) \;=\; \frac{d_k^{\log}(x,y)}{\bigl|\log_b b_k - \log_b a_k\bigr|},
\]
and convert it to a similarity in $[0,1]$ by
\[
s_k(x,y) \;=\; 1 - \mathrm{clip}_{[0,1]}\!\left(\tilde d_k^{\log}(x,y)\right).
\]

For the remaining keys (\texttt{lora\_dropout}, \texttt{num\_epochs}, \texttt{gradient\_accumulation\_steps}), we treat them as \emph{additive} and compare them linearly. For such a key $k$ with values $x=h_i[k]$, $y=h_j[k]$ and fixed range $[a_k,b_k]$, we define the \emph{linear normalized distance} as
\[
\tilde d_k^{\mathrm{lin}}(x,y) \;=\; \frac{|x-y|}{b_k-a_k},
\]
and again convert it to a similarity via
\[
s_k(x,y) \;=\; 1 - \mathrm{clip}_{[0,1]}\!\left(\tilde d_k^{\mathrm{lin}}(x,y)\right).
\]
In our experiments we use uniform weights across keys and define the hyperparameter similarity for the pair $(i,j)$ as the mean per-key similarity:
\[
s_{\mathrm{hp}}(i,j) \;=\; \frac{1}{|K|}\sum_{k\in K} s_k\!\bigl(h_i[k], h_j[k]\bigr),
\]
where $K$ is the set of keys listed above.

Finally, we summarize similarity across the five templates for a passage by averaging over all $\binom{5}{2}=10$ unordered pairs:
\[
S_{\mathrm{hp}} \;=\; \frac{1}{10}\sum_{1\le i<j\le 5} s_{\mathrm{hp}}(i,j).
\]

Intuitively, $S_{\text{text}}$ captures whether the model is repeatedly proposing the same passage transformation strategy, while $S_{\text{hp}}$ captures whether it is collapsing onto the same training recipe. Increasing similarity (especially in $S_{\text{text}}$) indicates reduced exploration and thus stronger mode collapse.

The raw numbers for intra-iteration similarity are available at the bottom of Tables \ref{tab:noarchive_templates} and \ref{tab:archive_evolution_templates}.

\subsection{Full numbers for average length of each self-edit's synthetic training data $D$}
\label{datalenfull}

\begin{table}[H]
\centering

\begin{minipage}[t]{0.49\textwidth}
  \centering
  {\scriptsize
    \setlength{\tabcolsep}{2pt}
    \renewcommand{\arraystretch}{1.12}
    \begin{tabularx}{\linewidth}{@{} >{\centering\arraybackslash}p{0.11\linewidth} Y Y Y Y @{}}
    \toprule
    \textbf{Iter.} &
    \textbf{SEAL Baseline, ``Implications''} &
    \textbf{SEAL Baseline, ``Rewrite''} &
    \textbf{Ours, ``No Archive''} &
    \textbf{Ours, ``With Archive''} \\
    \midrule
    0 & 2229.59 & 2409.31 & 2050.71 & 1470.26 \\
    1 & 2404.18 & 1891.73 & 1283.64 & 1564.20 \\
    2 & 2565.60 & 1798.61 &  925.94 & 1569.93 \\
    3 & 2975.01 & 1744.74 & 1133.71 & 1474.18 \\
    4 & 3194.05 & 1809.69 & 1191.90 & 1389.34 \\
    \bottomrule
    \end{tabularx}
  }
  \captionsetup{type=table,skip=2pt}
  \caption{Average number of characters in each self-edit's synthetic data on the 50-passage \textbf{training} subset.}
  \label{tab:train_synth_chars}
\end{minipage}\hfill
\begin{minipage}[t]{0.49\textwidth}
  \centering
  {\scriptsize
    \setlength{\tabcolsep}{2pt}
    \renewcommand{\arraystretch}{1.12}
    \begin{tabularx}{\linewidth}{@{} >{\centering\arraybackslash}p{0.11\linewidth} Y Y Y Y @{}}
    \toprule
    \textbf{Iter.} &
    \textbf{SEAL Baseline, ``Implications''} &
    \textbf{SEAL Baseline, ``Rewrite''} &
    \textbf{Ours, ``No Archive''} &
    \textbf{Ours, ``With Archive''} \\
    \midrule
    0 & 2262.75 & 2477.91 & 2915.83 & 1173.94 \\
    1 & 2487.64 & 1971.58 & 1198.09 & 1930.34 \\
    2 & 2632.56 & 1879.27 & 1755.12 & 1544.98 \\
    3 & 2989.57 & 1819.65 & 1417.14 & 1439.51 \\
    4 & 3200.19 & 1827.41 & 1286.89 & 1404.57 \\
    \bottomrule
    \end{tabularx}
  }
  \captionsetup{type=table,skip=2pt}
  \caption{Average number of characters in each self-edit's synthetic data on the 200-passage \textbf{validation} subset.}
  \label{tab:val_synth_chars}
\end{minipage}

\end{table}

\end{document}